\documentclass[a4paper]{cas-dc}

\usepackage[authoryear,longnamesfirst]{natbib}

\usepackage{subcaption}

\usepackage{mathrsfs}

\usepackage{xcolor}
\usepackage{bbding}
\usepackage{pifont}
\usepackage[linesnumbered,ruled]{algorithm2e}
\usepackage{caption}
\usepackage{placeins}

\def\tsc#1{\csdef{#1}{\textsc{\lowercase{#1}}\xspacet}}
\tsc{WGM}
\tsc{QE}

\begin{document}
\let\WriteBookmarks\relax
\def\floatpagepagefraction{1}
\def\textpagefraction{.001}

\shorttitle{}    

\shortauthors{}  

\newcommand{\MName}{HACMatch}
\newcommand{\augName}{PoseMosaic}

\title[mode = title]{\MName: Semi-Supervised Rotation Regression with Hardness-Aware Curriculum Pseudo Labeling}  



%

\author[1]{Mei Li}




\author[1]{Huayi Zhou}
\author[1]{Suizhi Huang}
\author[1]{Yuxiang Lu}
\author[1]{Yue Ding}
\author[1]{Hongtao Lu}
\ead{htlu@sjtu.edu.cn}

\cormark[1]


\affiliation[1]{organization={Shanghai Jiao Tong University},
            addressline={800 Dongchuan RD. Minhang District}, 
            city={Shanghai},
            postcode={200240}, 
            country={China}}







\cortext[1]{Corresponding author}



\begin{abstract}
Regressing 3D rotations of objects from 2D images is a crucial yet challenging task, with broad applications in autonomous driving, virtual reality, and robotic control. Existing rotation regression models often rely on large amounts of labeled data for training or require additional information beyond 2D images, such as point clouds or CAD models. Therefore, exploring semi-supervised rotation regression using only a limited number of labeled 2D images is highly valuable. While recent work FisherMatch introduces semi-supervised learning to rotation regression, it suffers from rigid entropy-based pseudo-label filtering that fails to effectively distinguish between reliable and unreliable unlabeled samples. To address this limitation, we propose a hardness-aware curriculum learning framework that dynamically selects pseudo-labeled samples based on their difficulty, progressing from easy to complex examples. We introduce both multi-stage and adaptive curriculum strategies to replace fixed-threshold filtering with more flexible, hardness-aware mechanisms. Additionally, we present a novel structured data augmentation strategy specifically tailored for rotation estimation, which assembles composite images from augmented patches to introduce feature diversity while preserving critical geometric integrity. Comprehensive experiments on PASCAL3D+ and ObjectNet3D demonstrate that our method outperforms existing supervised and semi-supervised baselines, particularly in low-data regimes, validating the effectiveness of our curriculum learning framework and structured augmentation approach.
\end{abstract}


    
    



\maketitle

\section{Introduction}

Rotation regression of objects from 2D images poses a fundamental challenge in computer vision, with applications in autonomous driving~\cite{auto1, auto2}, augmented reality~\cite{liu2023gen6d, he2023onepose, wen2024foundationpose}, and robotic control~\cite{10043016}. Accurate rotation regression is essential as it enables machines to better understand and interact with their environments, improving task performance and safety—critical factors for the advancement of intelligent systems. However, achieving high performance in rotation regression relies on large amounts of labeled data. Compared to tasks such as classification or dense prediction, obtaining accurate orientation annotations for objects is significantly more challenging, often requiring precise alignment of a mesh model with the images.

In order to reduce the dependence on large amounts of high-quality labeled data, recent research has shifted from traditional supervised learning to more flexible methods that leverage unlabeled data or auxiliary information. PoseContrast~\cite{xiao2021posecontrast} uses CAD models to assist in few-shot 3D rotation regression, while PA-Pose~\cite{LIU2024110151} incorporates point cloud information. However, these approaches either require data from modalities beyond 2D images or fail to fully exploit the vast availability of unlabeled image data.

A recent work, FisherMatch~\cite{fishermatch} starts to focus on rotation regression under semi-supervised learning. Before it, \cite{matFisher} introduces a method to directly assess the reliability of any matrix output by a regression network as a rotation matrix based on the matrix Fisher distribution. FixMatch~\cite{FixMatch}, a semi-supervised learning framework, enhances model performance by leveraging consistency regularization and confident pseudo-labeling of unlabeled data. Combining the Matrix Fisher Distribution and FixMatch, FisherMatch proposes a semi-supervised rotation regression framework that allows training with unlabeled data. However, FisherMatch suffers from its reliance on a fixed entropy threshold for pseudo-label filtering, which often fails to distinguish between reliable and unreliable unlabeled samples. This rigid approach limits the effectiveness of semi-supervised learning, especially in challenging scenarios, and underscores the need for a more adaptive solution.

To address this issue, we propose a curriculum learning framework for semi-supervised rotation regression. 
Our method selects pseudo-labeled samples based on their difficulty, starting with the easiest and gradually moving to harder ones, so the model learns step by step from confident, easier samples to more uncertain, complex ones.
We introduce both a \emph{multi-stage curriculum} and a lightweight \emph{adaptive curriculum} strategy. The former gradually increases the proportion of accepted samples based on entropy ranking across training stages, while the latter interpolates the entropy threshold throughout training. Both alleviate the rigidity of fixed-threshold filtering and lead to more effective use of unlabeled data.

Beyond the learning framework, the design of the strong data augmentation strategy is also important. While advanced techniques that create composite images from patches have proven powerful for classification tasks~\cite{yoco,cutmix}, they pose a dilemma for rotation regression. Such methods often disrupt the global object structure and holistic cues essential for accurate pose estimation. Furthermore, existing patch-based approaches typically apply variations of a single augmentation type across different patches (e.g., rotating each patch by a different angle). This limits the diversity of the learned features.

To address these challenges, we introduce {\augName}, a novel augmentation strategy with a two-part contribution. First, instead of varying the parameters of a single transformation, {\augName} creates a richer and more complex training signal by applying different types of augmentations to different patches of the same image, sampled from a diverse pool. Second, we recognize that not all augmentation types are suitable for this task. To mitigate the risk of structural corruption, we perform a systematic empirical study to select an optimal pool of augmentations, explicitly filtering out transformations that are detrimental to geometric fidelity (e.g., horizontal flipping). This dual approach ensures that {\augName} generates significant feature diversity at a local level while simultaneously preserving the global structural integrity crucial for robust rotation regression.

We conduct comprehensive experiments on the PASCAL3D+ and ObjectNet3D benchmarks to validate our approach. The results demonstrate that it outperforms existing supervised and semi-supervised baselines. The performance gains are particularly substantial in low-data regimes, highlighting the effectiveness of our method in leveraging unlabeled data. Furthermore, through extensive ablation studies, we systematically verify the individual contributions of our hardness-aware curriculum and the {\augName} augmentation, confirming that each component is essential to the final performance. We also show that {\augName} is superior to other advanced augmentation techniques like CutMix \cite{cutmix} and YOCO \cite{yoco}, validating its specific design for the rotation estimation task.

To summarize, our main contributions are as follows:
\begin{itemize}
    \item We propose HACMatch, a Hardness-Aware Curriculum Learning framework that replaces rigid filtering thresholds with a dynamic, hardness-aware mechanism for pseudo-label selection. We introduce and evaluate two effective implementations: a discrete Multi-stage strategy and a more lightweight continuous Adaptive strategy, both of which successfully guide the model from easy to complex examples.
    \item We introduce {\augName}, a novel data augmentation strategy that addresses the unique challenges of rotation regression. Its design is twofold: First, to maximize feature diversity, it applies different types of augmentations to different patches of an image, moving beyond prior methods that simply vary parameters of a single transformation. Second, to preserve essential geometric cues, the pool of available transformations is carefully selected through empirical analysis to discard structurally harmful operations.
    \item We conduct comprehensive experiments on PASCAL3D+ and ObjectNet3D to prove performance gains of our method, particularly in low-data regimes. We also rigorously validate the effectiveness and synergy of our proposed curriculum and augmentation components through extensive ablation studies.
\end{itemize}

\section{Related Work}
\subsection{Rotation Estimation}
Research in 3D rotation estimation is broadly categorized into theoretical work on representations \cite{matFisher, rotLaplace, mohlin2020probabilistic} and a progression of practical applications. Applications have evolved from instance-level \cite{rad2018bb8, tekin2018realtime, pavlakos20176dof}, to category-level \cite{9636212, 9933183}, and ultimately to estimating rotations for unknown objects \cite{pitteri2019cornet, gou2022unseen}.

\textbf{Rotation Representation.} A key challenge is the discontinuity of common representations like Euler angles. While rotation matrices can continuously represent the three-dimensional special orthogonal group (SO(3)), ensuring a network's output is a valid SO(3) matrix is non-trivial \cite{mohlin2020probabilistic}. To address this, \cite{matFisher} introduced a loss based on the matrix Fisher distribution to enforce valid rotations, which \cite{rotLaplace} later improved with a Laplace distribution for long-tailed data. Other works have explored further optimizations, such as using implicit representations for uncertainty \cite{murphy2022implicitpdf} or designing SO(3) and SO(2)-equivariant networks \cite{klee2023image, howell2023equivariant, liu2023delving}.

\textbf{Rotation Estimation Applications.} Application-focused work has progressed from instance-level \cite{rad2018bb8, tekin2018realtime, pavlakos20176dof} and category-level estimation \cite{9636212, 9933183} to handling unseen object categories \cite{pitteri2019cornet, gou2022unseen}. More recent studies also investigate the application of few-shot learning \cite{xiao2022fewshot} and performance in real-world scenarios \cite{fu2022categorylevel}.

\subsection{Curriculum Learning}
Inspired by human learning, curriculum learning (CL) trains models by progressing from easier to harder data \cite{bengio2012evolving, wang2021survey}. A standard CL framework includes a Difficulty Measurer to assess sample complexity and a Training Scheduler to organize the learning process based on difficulty \cite{wang2021survey}.

Self-paced Learning (SPL), where the model assesses data difficulty via training loss, is a form of CL. In semi-supervised learning (SSL), CL can enhance pseudo-labeling techniques. For example, Curriculum Labeling \cite{CLSemi} uses a self-paced strategy to refine pseudo-labels, while Curriculum Pseudo Labeling (CPL) \cite{zhang2022flexmatch} dynamically adjusts thresholds to better leverage unlabeled data.

\begin{figure*}[!tbp]
    \centering
    \includegraphics[width=\textwidth]{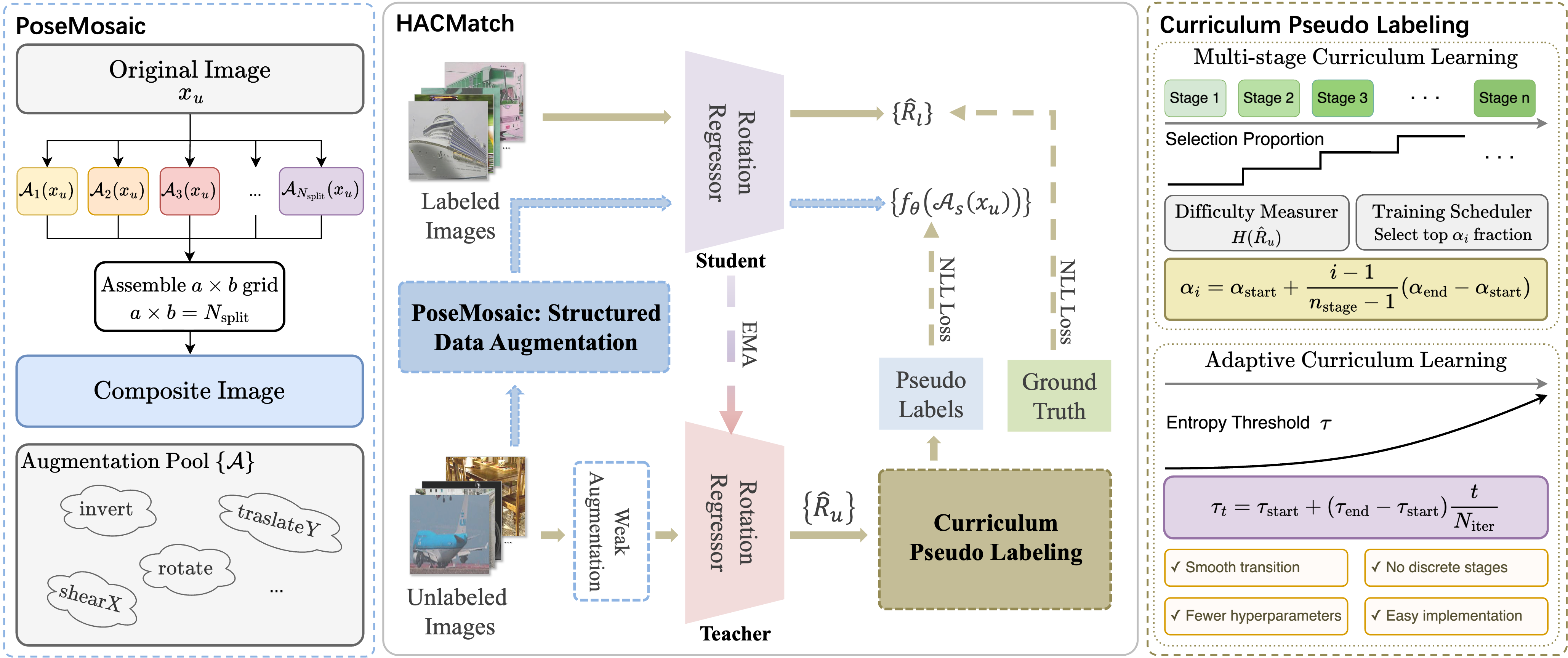}
    \caption{\textbf{The overall framework (HACMatch).} (Middle) Semi-supervised Pipeline: The framework operates in two stages: supervised pre-training with labeled images, followed by semi-supervised learning where strongly augmented unlabeled images are fed to the student network while weakly augmented versions go to the teacher network. Curriculum pseudo-labeling selects reliable pseudo-labels $\{\hat{R}_u\}$ based on prediction entropy. \textbf{(Left) Structured Data Augmentation (PoseMosaic):} Given an original image $x_u$, we apply $n$ different augmentation transformations $\{\mathcal{A}_i\}$ to generate augmented views $\{\tilde{x}_u^{(i)}\}$. These views are spatially assembled into an $a \times b$ grid where $a \cdot b = n$, creating a composite image that preserves global structure while enhancing local diversity. \textbf{(Right) Curriculum Learning Strategies:} We propose two approaches: \textit{Multi-stage Curriculum Learning} divides training into discrete stages with linearly increasing selection proportions using entropy $H(\hat{R}_u)$; \textit{Adaptive Curriculum Learning} continuously adjusts the entropy threshold throughout training, offering smoother transitions.}
    \label{fig:pipeline}
\end{figure*}

\subsection{Semi-supervised Learning}
Semi-supervised learning (SSL) methods are principally divided into four categories \cite{DeepSemiSuvey}. Generative methods leverage models like Generative Adversarial Networks (GANs) \cite{goodfellow2014generative} and Variational Autoencoders (VAEs) \cite{kingma2022autoencoding} to learn the data distribution and create new samples \cite{odena2016semisupervised, radford2016unsupervised}. Consistency regularization methods ensure that model outputs remain invariant to data perturbations \cite{park2017adversarial, xie2020unsupervised}. Graph-based methods construct a data similarity graph to propagate labels \cite{iscen2019label, 9157594, li2020densityaware}. Pseudo-labeling uses a model's own high-confidence predictions on unlabeled data as training labels \cite{trinet}. Furthermore, hybrid methods like FixMatch \cite{FixMatch} combine techniques such as pseudo-labeling with consistency regularization to boost performance.

{To further address the issue of noisy or unreliable pseudo-labels, especially in the early training stages, recent advances have explored pseudo-label alignment and refinement mechanisms~\cite{zheng2021rectifying, hu2023pseudo}. For instance, PAIS~\cite{hu2023pseudo} proposed a dynamic aligning framework to reduce pseudo-label noise and improve cross-view consistency in semi-supervised instance segmentation. }

\section{Method}

Our work focuses on regressing the object's rotation from its 2D image without data of any other modality in a semi-supervised learning (SSL) setting. 
The core idea of our model consists of two main components: hardness-aware curriculum learning (Section~\ref{sec:cl}) and a structured image augmentation strategy termed {\augName} (Section~\ref{sec:aug}) specifically designed for rotation estimation. The curriculum learning determines which samples are eligible to participate in the semi-supervised training process and at what stage they are introduced. The {\augName} aims to improve the SSL model's performance by enhancing data diversity, which applies different types of augmentations to different pathces of an image.

\subsection{Preliminary}
The training dataset $\mathcal{D}$ of semi-supervised rotation estimation is composed of a small set of labeled data $\mathcal{D}_l = \{(x_{l_i}, R_{l_i})\}_{i=1}^{N_l}$ and a large set of unlabeled data $\mathcal{D}_u = \{x_{u_i}\}_{i=1}^{N_u}$, where $N_l \ll N_u$. For each labeled pair $(x_l, R_l)$, $x_l$ is an input image and $R_l \in SO(3)$ is the corresponding ground-truth $3 \times 3$ rotation matrix. The model, denoted as a student network $f_\theta$ with trainable parameters $\theta$, predicts a rotation matrix $\hat{R} = f_\theta(x)$. The training on the labeled set is supervised by loss: 
\begin{equation}
    \mathcal{L}_s = \frac{1}{|\mathcal{D}_l|} \sum_{x_l \in \mathcal{D}_l}\mathcal{L}(f_\theta(x_l), R_{l}),
\end{equation}
where $\mathcal{L}$ is the negative log likelihood (NLL) loss in FisherMatch~\cite{fishermatch}.

\begin{figure*}[!tbp]
    \centering
    \begin{subfigure}{0.35\textwidth}
        \centering
        \includegraphics[width=\linewidth]{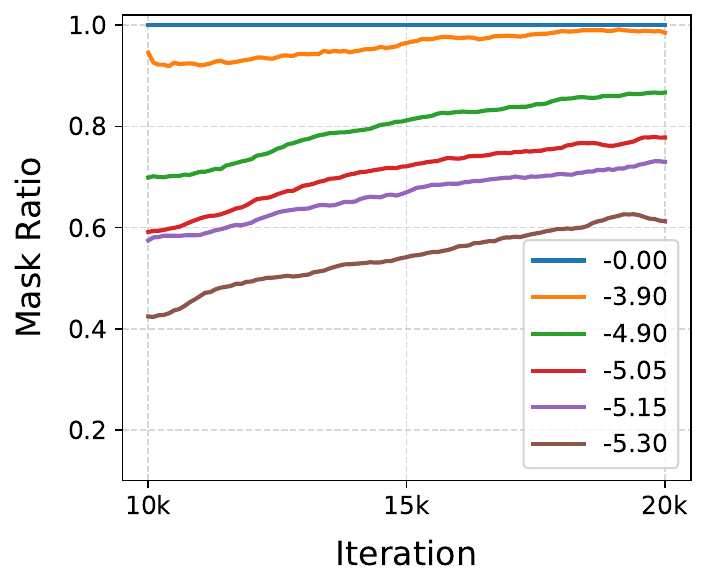}
        \caption{mask ratio vs $\tau$.}
        \label{fig:mask_ratio}
    \end{subfigure} \hspace{10mm}  
    \begin{subfigure}{0.35\textwidth}
        \centering
        \includegraphics[width=\linewidth]{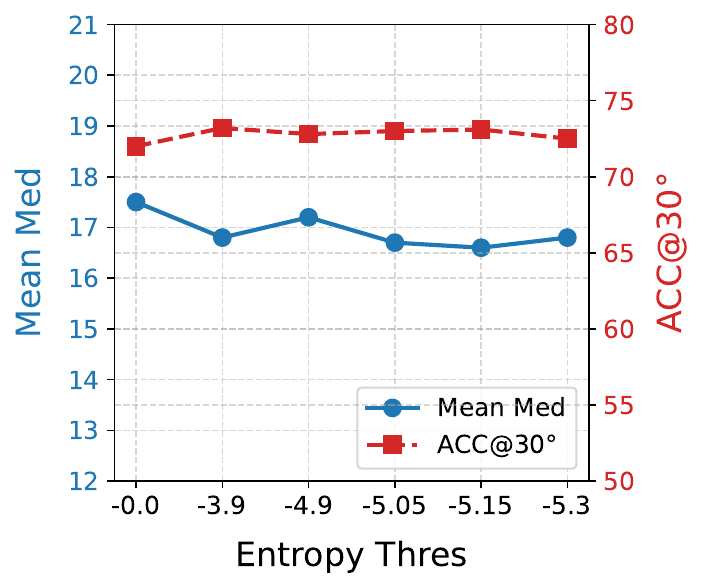}
        \caption{performance vs $\tau$ .}
        \label{fig:entropy_perform}
    \end{subfigure}
    \caption{\textbf{The illustration of limitation of using a fixed SSL filtering threshold $\tau$.} \(\tau=-0.0\) means entropy-based filtering does not work. (a) The evolution of the Mask Ratio (the proportion of unlabeled samples in each training iteration whose entropy below \(\tau\)) for different fixed entropy thresholds (\(\tau\)).  For any given \(\tau\), the ratio stays within a narrow band. (b) Final model performance (Mean Med and ACC@30$^\circ$) across different \(\tau\) values. Performance is insensitive to the choice of the fixed threshold.}
    \label{fig:entropy_thres}
\end{figure*}

To leverage the vast amount of unlabeled data, we adopt a consistency regularization framework based on a student-teacher paradigm~\cite{meanteacher, FixMatch}. Semi-supervised training methods based on the consistency principle generate pseudo-labels for unlabeled samples using a teacher model, $f_{\theta'}$. The teacher model has the same architecture as the student model, and its weights $\theta'$ are typically an exponential moving average (EMA) of the student's weights $\theta$. For an unlabeled image $x_u$, a pseudo-label $\hat{R}_u$ is generated by the teacher network using a weakly-augmented version of the image, i.e., $\hat{R}_u = f_{\theta'}(\mathcal{A}_w(x_u))$, where $\mathcal{A}_w(\cdot)$ is a weak augmentation function. A entropy estimation function $H(\cdot)$, which is same as defined in FisherMatch~\cite{fishermatch}, is then used to assess the reliability of this pseudo-label. Only pseudo-labels with an entropy lower than a predefined threshold $\tau$ are retained for training. The student network is then trained to produce a consistent prediction on a strongly-augmented version of the same image, $\mathcal{A}_s(x_u)$. This consistency loss is formulated as:
\begin{equation}
    \mathcal{L}_u = \frac{1}{|\mathcal{D}_u|} \sum_{x_u \in \mathcal{D}_u} \mathbb{1}\big(H(\hat{R}_u) \leq \tau \big) \mathcal{L}\big(f_\theta(\mathcal{A}_s(x_u)), \hat{R}_u\big),
\end{equation}
where $\mathbb{1}(\cdot)$ is the indicator function. The final training objective is a combination of the supervised and unsupervised losses:
\begin{equation}
    \mathcal{L}_{\text{total}} = \mathcal{L}_s + \lambda \mathcal{L}_u,
\end{equation}
where $\lambda$ is a scalar hyperparameter that balances the contribution of the two loss terms.

\subsection{Hardness-aware Curriculum Pseudo Labeling} \label{sec:cl}

\subsubsection{Limitations of a Fixed Entropy Threshold}

In traditional consistency-based semi-supervised learning, the pseudo-label filter threshold \(\tau\) plays a pivotal role. It governs the pseudo-label filtering process, determining which unlabeled samples are considered reliable enough to be used in training the student network. We define the "Mask Ratio" as the proportion of unlabeled samples in each training iteration whose pseudo-labels meet the confidence criterion (i.e., entropy below \(\tau\)).

Ideally, the training process should follow a curriculum where the model learns from progressively more challenging examples. This implies that the Mask Ratio should start low, ensuring that only the most confident pseudo-labels are used in the early stages, and then gradually increase as the model's predictive accuracy improves. Such a dynamic would allow the model to leverage an ever-increasing amount of unlabeled data.

However, our experiments indicate that a fixed threshold \(\tau\) is inherently limited in facilitating this desired behavior. As shown in Figure~\ref{fig:mask_ratio}, for any single choice of \(\tau\), the corresponding Mask Ratio evolves within a very narrow range throughout the training process. For example, a more relaxed threshold (e.g., \(\tau = -3.90\)) results in a consistently high Mask Ratio (above 90\%), failing to effectively filter out potentially noisy pseudo-labels at the beginning of training. Conversely, a stricter threshold (e.g., \(\tau = -5.30\)) confines the Mask Ratio to a low level, preventing the model from utilizing a larger fraction of the unlabeled data even after it has become more proficient. This rigidity demonstrates that a fixed \(\tau\) cannot adapt to the model's evolving competence.

Furthermore, simply tuning \(\tau\) to a different fixed value does not resolve the core issue or significantly impact final performance. Figure~\ref{fig:entropy_perform} shows that despite the wide variation in Mask Ratios resulting from different thresholds (ranging from approximately 43\% to nearly 95\%), the key performance metrics of Mean Med and Accuracy@30$^\circ$ remain stable. To overcome this limitation and fully exploit the unlabeled dataset, a more dynamic approach is required to control how samples are incorporated into the semi-supervised training process.

\begin{table*}[ht]
    \centering
    \caption{Comparison of Method Implementation Details for Multi-stage Curriculum Learning and Adaptive Curriculum Learning.}
    \begin{tabular}{l|l|l}
    \toprule
        \textbf{Comparison Aspect} & \textbf{Multi-stage CL} & \textbf{Adaptive CL} \\ \midrule
        Hyper-parameters & 3 parameters & 2 parameters \\
        & ($\alpha_{\text{start}}$, $\alpha_{\text{end}}$, $n_{stage}$) & ($\tau_{\text{start}}$, $\tau_{\text{end}}$) \\ \midrule
        Core Mechanism & Adjusts the \textbf{selection proportion} ($\alpha_i$) & Directly adjusts the \textbf{entropy threshold} ($\tau_t$) \\
        & of samples per batch. & for filtering. \\ \midrule
        Adjustment Schedule & \textbf{Discrete, stepped} & \textbf{Continuous, linear} \\
        & Proportion $\alpha_i$ is fixed within each stage. & Threshold $\tau_t$ is updated every iteration. \\ \midrule
        Threshold Calculation & \textbf{Batch-dependent} & \textbf{Batch-independent} \\
        & Threshold $z_i$ is the $\alpha_i$-quantile & Threshold $\tau_t$ is calculated based on \\
        & of the current batch's entropies. & the global iteration step $t$. \\ \bottomrule
    \end{tabular}
    \label{tab:cl_comparison_en}
\end{table*}


\subsubsection{Multi-stage Curriculum Learning}
To address the inflexibility of a fixed threshold, we first introduce a multi-stage curriculum learning mechanism. This approach is built upon two core components: a \textit{difficulty measurer}, which quantifies the reliability of a pseudo-label, and a \textit{training scheduler}, which controls the introduction of unlabeled samples into the training process based on their measured difficulty.

For rotation estimation, we use the entropy of the predicted Fisher distribution~\cite{fishermatch}, denoted as $H(\hat{R}_t)$, as the difficulty measurer, where a lower entropy signifies a more confident (i.e., "easier") prediction from the teacher model $f_{\theta'}$. The training scheduler leverages this measure to progressively introduce more challenging samples as the student model $f_\theta$ becomes more proficient.

\begin{algorithm}[t]
\caption{Stage-Specific Threshold Calculation for Multi-stage CL}
\label{alg:multistage_cl}
\KwIn{Current iteration $t$, total iterations $T_{total}$, number of stages $n_{\text{stage}}$, start proportion $\alpha_{\text{start}}$, end proportion $\alpha_{\text{end}}$, batch of teacher predictions' entropies $\{H(\hat{R}_t^{(j)})\}_{j=1}^N$}
\KwOut{Stage-specific entropy threshold $z_i$ for the current iteration}

Compute stage length: $L_{\text{stage}} \gets \left\lfloor N_{iter} / n_{\text{stage}} \right\rfloor$ \;
Determine current stage index: $i \gets \min\left(\left\lfloor t / L_{\text{stage}} \right\rfloor + 1, n_{\text{stage}}\right)$ \;
Compute stage proportion: 
\[\alpha_i \gets \alpha_{\text{start}} + \frac{i - 1}{n_{\text{stage}} - 1} (\alpha_{\text{end}} - \alpha_{\text{start}})\]

Sort $\{H(\hat{R}_t^{(j)}\}$ in ascending order \;
Compute quantile index: $k \gets \text{floor}(\alpha_i |\{H(\hat{R}_t^{(j)}\}|)$ \;
Set stage-specific threshold: $z_i \gets \{H(\hat{R}_t^{(j)}\}[k]$ \;

\Return{$z_i$} \;
\end{algorithm}

The scheduler operates by dividing the total number of semi-supervised training iterations, $N_{iter}$, into $n_{stage}$ distinct stages of equal length. For each stage $i \in \{1, 2, \ldots, n_{stage}\}$, we define a selection proportion $\alpha_i$, which increases linearly from a starting value $\alpha_{\text{start}}$ to an ending value $\alpha_{\text{end}}$:
\begin{equation}
    \alpha_i = \alpha_{\text{start}} + \frac{i - 1}{n_{\text{stage}} - 1} (\alpha_{\text{end}} - \alpha_{\text{start}}).
\end{equation}

In any given training batch within stage $i$, we select the top $\alpha_i$ fraction with the lowest entropy values. Let $z_i$ be the entropy value of the sample at the $\alpha_i$-quantile within this filtered pool (Algorithm \ref{alg:multistage_cl}). The unsupervised loss for an unlabeled sample $x_u$ with teacher prediction $\hat{R}_t$ and student prediction $\hat{R}_s$ is then defined as:
\begin{equation}
\mathcal{L}_u(x_u) = \frac{1}{|\mathcal{D}_u|}\mathbb{1}\left(H(\hat{R}_t) \leq z_i \right) \mathcal{L}(\hat{R}_s, \hat{R}_t).
\end{equation}
Here, $z_i$ serves as a dynamic, stage-specific threshold that becomes more relaxed over time. In this framework, entropy acts dually as a preliminary filter and as a fine-grained difficulty measurer that guides the curriculum.

\subsubsection{Adaptive Curriculum Learning}
While our multi-stage CL strategy provides a structured curriculum, it introduces several new hyperparameters that require careful tuning: the start proportion $\alpha_{\text{start}}$, the end proportion $\alpha_{\text{end}}$, and the number of stages $n_{stage}$. To mitigate this complexity, we also propose a simpler, yet effective, \textit{adaptive curriculum learning strategy}.

\begin{figure*}[!htbp]
    \centering
    \begin{subfigure}{0.24\textwidth}
        \includegraphics[width=\textwidth]{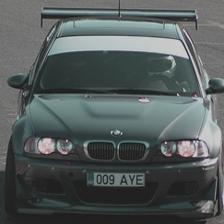}
        \caption{Original}
    \end{subfigure}
    \begin{subfigure}{0.24\textwidth}
        \includegraphics[width=\textwidth]{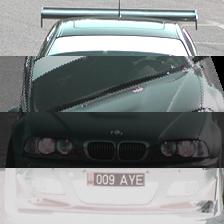}
        \caption{(4, 1)}
    \end{subfigure} 
    \begin{subfigure}{0.24\textwidth}
        \includegraphics[width=\textwidth]{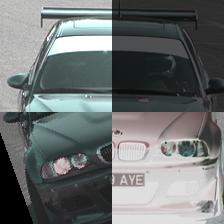}
        \caption{(2, 2)}
        \label{fig:rot_exp_3}
    \end{subfigure}
    \begin{subfigure}{0.24\textwidth}
        \includegraphics[width=\textwidth]{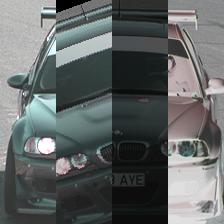}
        \caption{(1, 4)}
    \end{subfigure}
    \caption{Visualization of our proposed \textbf{{\augName}} method with $n=4$. The original image (a) is reassembled from four independently augmented versions into grids of (b) 4$\times$1, (c) 2$\times$2, and (d) 1$\times$4 patches.}
    \label{fig:rotationsoup_example}
\end{figure*}

\begin{algorithm}[!tbp]
\caption{{\augName}}
\label{alg:rotation-soup}
\KwIn{$\text{img} \in \mathbb{R}^{W \times H \times C}$: Input image, $n$: Number of patches}
\KwOut{Augmented image with mixed patches}

Initialize empty list $\texttt{split\_list} \leftarrow []$\;
\For{$i \leftarrow 1$ \KwTo $n$}{
    \If{$n \bmod i = 0$}{
        $\texttt{split\_list}.\texttt{append}((i, n // i))$\;
    }
}
Randomly select $(a, b)$ from $\texttt{split\_list}$\;
Initialize $\texttt{ret\_img} \leftarrow \texttt{img}$\;
\For{$i \leftarrow 0$ \KwTo $n-1$}{
    $p \leftarrow i \bmod b$\;
    $q \leftarrow i // b$\;
    
    $\texttt{aug\_img} \leftarrow \mathcal{A}_i(\texttt{img}) \quad \mathcal{A}_i \in \{\mathcal{A}$\}\;

    $\texttt{ret\_img}[p \cdot \tfrac{W}{b} : (p+1) \cdot \tfrac{W}{b},\; q \cdot \tfrac{H}{a} : (q+1) \cdot \tfrac{H}{a},\; :] \leftarrow$

    \hspace{3em}$\texttt{aug\_img}[p \cdot \tfrac{W}{b} : (p+1) \cdot \tfrac{W}{b},\; q \cdot \tfrac{H}{a} : (q+1) \cdot \tfrac{H}{a},\; :]$\;
}
\Return $\texttt{ret\_img}$\;
\end{algorithm}

This approach eliminates discrete stages and instead dynamically adjusts the entropy threshold $\tau$ itself throughout training. We achieve this by interpolating the threshold value from a starting threshold $\tau_{\text{start}}$ to an ending threshold $\tau_{\text{end}}$ over the course of the semi-supervised training phase. Specifically, at any given iteration $t$, the adaptive threshold $\tau_t$ is computed as:
\begin{equation}
\tau_t = \tau_{\text{start}} + (\tau_{\text{end}} - \tau_{\text{start}}) \cdot \frac{t}{N_{iter}},
\end{equation}
where $N_{iter}$ is the total number of iterations allocated for semi-supervised training.

This method directly manipulates the filtering criterion in a smooth and continuous manner, obviating the need for staging and quantile-based selection within batches. Compared to the multi-stage strategy, the adaptive approach streamlines the implementation of dynamic sample selection and reduces the hyperparameter tuning effort, requiring only the definition of a start and an end threshold.

\subsubsection{Method Comparison}
Table \ref{tab:cl_comparison_en} conclude the practical differences between our two proposed curriculum strategies. The primary distinction lies in their control variable and update mechanism. The multi-stage CL approach operates on a discrete, stepped schedule, adjusting the \textit{selection proportion} ($\alpha_i$) at fixed intervals (stages). This requires a dynamic, batch-dependent calculation to find the corresponding entropy quantile ($z_i$) that serves as the effective threshold. In contrast, the adaptive CL strategy is continuous and simpler, directly interpolating the \textit{entropy threshold} ($\tau_t$) itself. This threshold is calculated purely as a function of the global training iteration $t$, making it batch-independent and eliminating the complexity of managing stages and in-batch sorting. 

\subsection{{\augName}} \label{sec:aug}

In our semi-supervised framework, consistency regularization requires a strong augmentation applied to the student's input. While advanced methods that create composite images from augmented views, such as YOCO~\cite{yoco}, have proven effective for classification, this strategy presents a critical dilemma for rotation estimation. On one hand, rotation regression is fundamentally dependent on preserving the object’s global structure and holistic shape cues, which can be easily corrupted by naively partitioning and reassembling an image. On the other hand, existing composite methods often limit feature diversity by applying variations of a single augmentation type across all patches. This motivates the need for an augmentation strategy that can resolve this tension.

\subsubsection{The Design of \augName}
We propose a novel data augmentation method for semi-supervised rotation estimation, termed \textbf{{\augName}}, which enhances model performance in this task (Algorithm \ref{alg:rotation-soup}). Specifically, given an unlabeled image $x_u \in \mathbb{R}^{W \times H \times C}$, we randomly apply $n$ different augmentation functions to generate a set of augmented images:
\begin{equation}
    \{ \tilde{x}_u^{(i)} = \mathcal{T}_i(x_u) \mid i = 1, 2, \ldots, n \},\quad \mathcal{A}_i \in \{\mathcal{A}\},
\end{equation}
where $\{\mathcal{A}\}$ is the predefined augmentation pool (e.g., color jitter, crop, blur, rotation), and each $\mathcal{A}_i$ is independently sampled. These $n$ images are then spatially partitioned and assembled into a new image of the same size as the original input by splitting the $W \times H$ spatial domain into $a \times b$ uniform sub-regions, where $(a,b)$ is randomly selected from the set $\{(u,v)|u, v\in N \text{ and } u\times v=n \}$. The composed image can be represented as:
\begin{equation}
    \hat{x}_u = \texttt{Assemble}(\{\tilde{x}_u^{(i)}\}_{i=1}^{n},\; a, b),\quad \hat{x}_u \in \mathbb{R}^{W \times H \times C}.
\end{equation}

\begin{figure*}[!htbp]
    \centering

    \begin{subfigure}[b]{0.12\textwidth}
        \includegraphics[width=\textwidth]{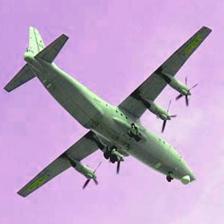}
        \caption{Contrast}
        \label{fig:yoco-contrast}
    \end{subfigure}
    \begin{subfigure}[b]{0.12\textwidth}
        \includegraphics[width=\textwidth]{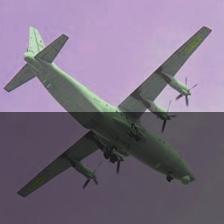}
        \caption{Brightness}
        \label{fig:yoco-brightness}
    \end{subfigure}
    \begin{subfigure}[b]{0.12\textwidth}
        \includegraphics[width=\textwidth]{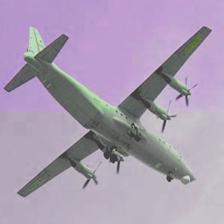}
        \caption{Color}
    \end{subfigure}
    \begin{subfigure}[b]{0.12\textwidth}
        \includegraphics[width=\textwidth]{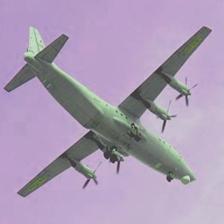}
        \caption{Cutout}
        \label{fig:vis_cutout}
    \end{subfigure}
    \begin{subfigure}[b]{0.12\textwidth}
        \includegraphics[width=\textwidth]{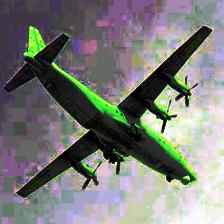}
        \caption{Equalize}
        \label{fig:yoco-equalize}
    \end{subfigure}
    \begin{subfigure}[b]{0.12\textwidth}
        \includegraphics[width=\textwidth]{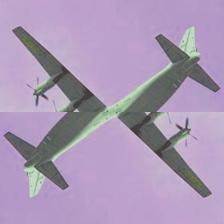}
        \caption{Flip}
        \label{fig:yoco-flip}
    \end{subfigure}
    \begin{subfigure}[b]{0.12\textwidth}
        \includegraphics[width=\textwidth]{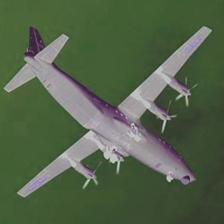}
        \caption{Invert}
        \label{fig:yoco-invert}
    \end{subfigure}    
    \begin{subfigure}[b]{0.12\textwidth}
        \includegraphics[width=\textwidth]{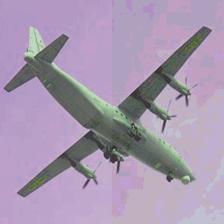}
        \caption{Posterize}
    \end{subfigure}

    \vspace{0.5em}
    
    \begin{subfigure}[b]{0.12\textwidth}
        \includegraphics[width=\textwidth]{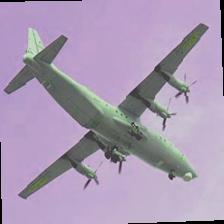}
        \caption{Rotate}
    \end{subfigure}
    \begin{subfigure}[b]{0.12\textwidth}
        \includegraphics[width=\textwidth]{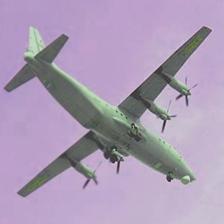}
        \caption{Sharpness}
    \end{subfigure}
    \begin{subfigure}[b]{0.12\textwidth}
        \includegraphics[width=\textwidth]{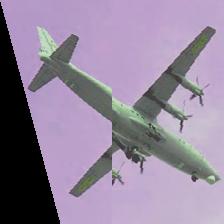}
        \caption{ShearX}
    \end{subfigure}
    \begin{subfigure}[b]{0.12\textwidth}
        \includegraphics[width=\textwidth]{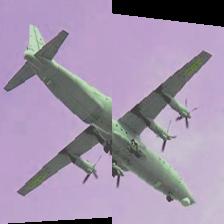}
        \caption{shearY}
    \end{subfigure}
    \begin{subfigure}[b]{0.12\textwidth}
        \includegraphics[width=\textwidth]{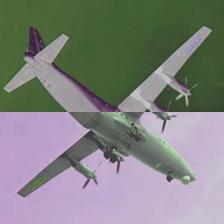}
        \caption{solarize}
    \end{subfigure}
    \begin{subfigure}[b]{0.12\textwidth}
        \includegraphics[width=\textwidth]{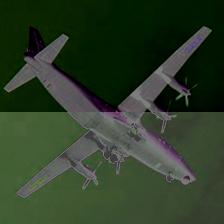}
        \caption{solarizeAdd}
    \end{subfigure}
    \begin{subfigure}[b]{0.12\textwidth}
        \includegraphics[width=\textwidth]{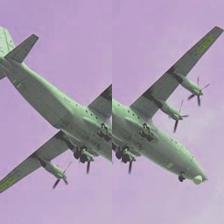}
        \caption{translateX}
    \end{subfigure}
    \begin{subfigure}[b]{0.12\textwidth}
        \includegraphics[width=\textwidth]{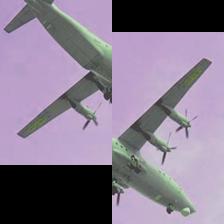}
        \caption{translateY}
    \end{subfigure}
    \caption{Visualization of various image augmentation methods. Each technique introduces different visual perturbations, with varying impacts on the task of rotation estimation.}
    \label{fig:aug_visualization}
\end{figure*}

\begin{figure*}[!htbp]
    \centering
    \includegraphics[width=\linewidth]{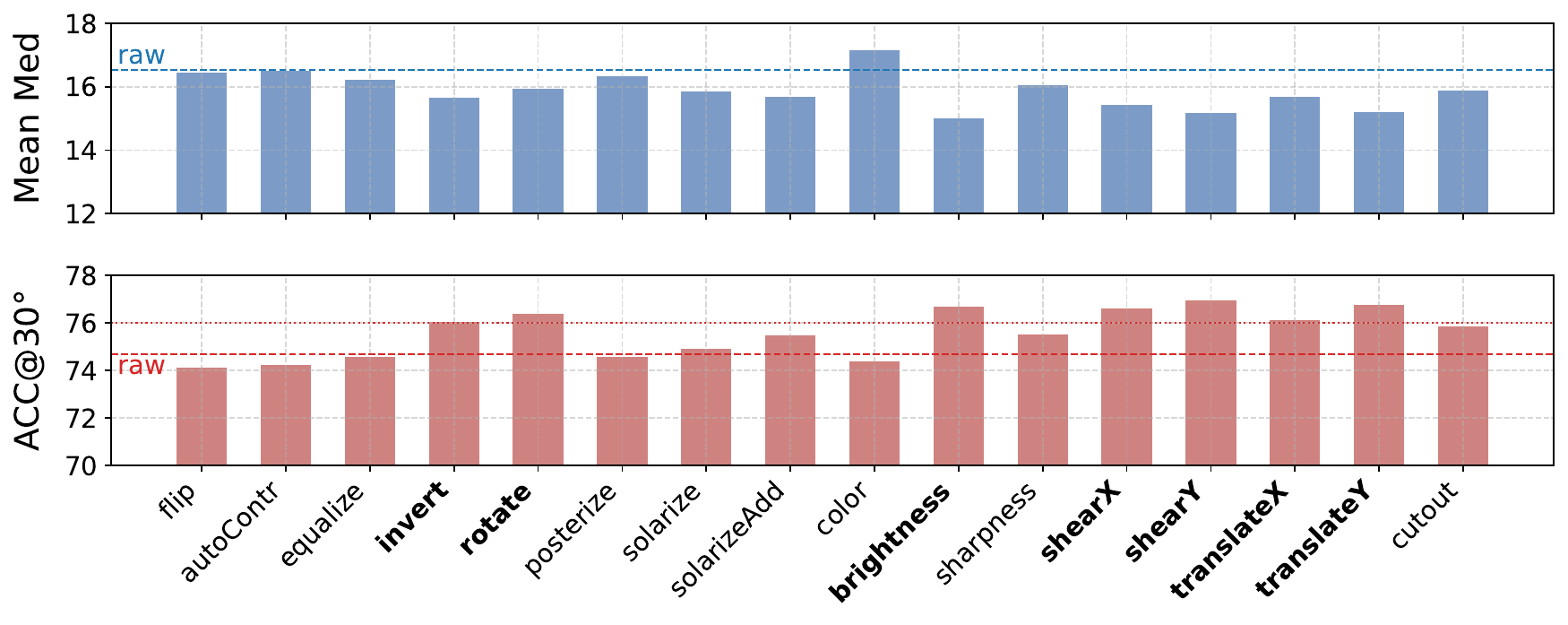}
    \caption{Performance comparison of 16 different augmentation strategies in a semi-supervised setting (PASCAL3D+, 5\% labeled data). The dashed lines represent the baseline (`raw') performance. We selected methods with ACC@30$^\circ$ > 76\% (indicated by \textbf{bold} labels) for our final augmentation pool.}
    \label{fig:aug_ablation}
\end{figure*}

Each sub-region of $\hat{x}_u$ corresponds to a patch from an independently augmented version of $x_u$, encouraging spatially diverse feature learning. This process can be interpreted as a structured, spatial augmentation strategy that blends local transformations while preserving global semantic alignment.
Figure~\ref{fig:rotationsoup_example} provides a visualization for $n=4$, where the original image (a) could be transformed into composite images with (4, 1), (2, 2), and (1, 4) patch arrangements, and one of the three augmented images will be fed to the student model during training.

{Since an object's orientation relates to its shape rather than its color or surface appearance, this augmentation technique preserves the essential outline needed for orientation identification. Although the augmented image resembles a concatenation (or mosaic) of different patches, the model can still determine the orientation as long as the object's overall structure is maintained. For example, in Figure \ref{fig:rot_exp_3}, the car is composed of four different parts. The top two patches and the bottom-right patch differ from the original image only in color, while the bottom-left patch is slightly rotated. Thus, it remains easy to judge that the car is facing forward.}

In our semi-supervised framework, {\augName} serves as the strong augmentation function $\mathcal{A}_s(\cdot)$ applied to unlabeled images for the student network. The teacher network, in contrast, receives a standard, weakly augmented version $\mathcal{A}_w(\cdot)$ of the original image. The consistency loss then compels the student model to learn rotation-invariant features from the complex, patched-together views, using the teacher's prediction on a cleaner view as guidance.

\subsubsection{Augmentation Pool Selection}
{Not every augmentation method is suitable for the pool $\{\mathcal{A}\}$ of {\augName}, as orientation estimation requires preserving the object's structural integrity. In fact, some augmentations can even be detrimental to the final performance.} The effectiveness depends critically on the quality of the underlying augmentation functions in $\{\mathcal{A}\}$. To identify the most beneficial transformations, we test 16 individual augmentation methods, each applied in a simple patches number $n=2$ configuration on the PASCAL3D+ dataset using 5\% of labeled data. The performance of each method is reported in Figure~\ref{fig:aug_ablation}, with the dashed lines noted as `raw' indicating the baseline performance without any strong augmentation.

\begin{table*}[!htbp]
    \centering
    \caption{The average 3D rotation regression performance (\textbf{Mean Med($^{\circ}$)$\downarrow$/ACC@30$(\%)^{\circ}$$\uparrow$}) across all 12 categories on \textbf{PASCAL3D+} \cite{pascal} under different ratios of labeled training data. Each value is reported as the mean $\pm$ standard deviation, computed over three independent runs. The best results are marked as \textbf{bold}.}
    \begin{tabular}{l|cccc}
    \toprule
        \multirow{2}{*}{Method} & \multicolumn{4}{c}{Ratio of Labled Samples} \\ 
        \cline{2-5}
        ~ & 5\% & 10\% & 20\% & All \\  
        \midrule
        Sup.-Fisher & 37.43{\scriptsize $\pm$0.57}/50.22{\scriptsize $\pm$0.62} & 32.97{\scriptsize $\pm$0.35}/55.87{\scriptsize $\pm$0.29} & 24.09{\scriptsize $\pm$0.42}/63.38{\scriptsize $\pm$0.65} & 12.63{\scriptsize $\pm$0.55}/83.76{\scriptsize $\pm$0.76} \\ 
        Sup.-Laplace & 36.29{\scriptsize $\pm$0.55}/52.73{\scriptsize $\pm$0.68} & 28.03{\scriptsize $\pm$0.51}/60.13{\scriptsize $\pm$0.75} & 22.66{\scriptsize $\pm$0.64}/65.38{\scriptsize $\pm$0.57} & 11.21{\scriptsize $\pm$0.74}/84.22{\scriptsize $\pm$0.62} \\   
        FisherMatch & 16.48{\scriptsize $\pm$0.59}/74.73{\scriptsize $\pm$0.82} & 13.65{\scriptsize $\pm$0.56}/79.13{\scriptsize $\pm$0.64} & 11.42{\scriptsize $\pm$0.37}/84.07{\scriptsize $\pm$0.58} & --/-- \\ 
        Ours(multi-stage) & 14.29{\scriptsize $\pm$0.44}/78.90{\scriptsize $\pm$0.74} & \textbf{12.26}{\scriptsize $\pm$0.49}/82.68{\scriptsize $\pm$0.56} & \textbf{10.73}{\scriptsize $\pm$0.68}/85.19{\scriptsize $\pm$0.47} & --/-- \\ 
        Ours(adaptive) & \textbf{14.17}{\scriptsize $\pm$0.36}/\textbf{79.24}{\scriptsize $\pm$0.61} & 12.37{\scriptsize $\pm$0.43}/\textbf{83.25}{\scriptsize $\pm$0.57} & 10.78{\scriptsize $\pm$0.52}/\textbf{85.69}{\scriptsize $\pm$0.66} & --/-- \\
    \bottomrule
    \end{tabular}
    \label{tab:res-pascal}
\end{table*}

The results show a non-trivial variance in impact. Some augmentations, such as invert, rotate, and brightness, yield substantial improvements over the baseline. Conversely, others like \textit{equalize} and \textit{flip} are detrimental, degrading model accuracy. Based on this analysis, we select the final augmentation pool $\{\mathcal{A}\}$ to include all methods that achieved an ACC@30$^{\circ}$ score greater than 76\%. These selected, high-performing methods are highlighted with \textbf{bolded} labels on the x-axis of Figure~\ref{fig:aug_ablation}.

Figure~\ref{fig:aug_visualization} provides a visualization of these diverse augmentation techniques. The selected methods generally preserve key structural features. For example, invert reverses the pixel values, brightness modifies luminance, shearX/Y applies a linear shear transformation, and translateX/Y shifts the image along an axis. {However, some augmentations are ineffective or even detrimental, like images \ref{fig:vis_cutout}(Cutout) and \ref{fig:yoco-flip}(Flip).
The effectiveness of {\augName} depends on appropriate pool selection, we will discuss this in Section \ref{sec:abl_aug}  in detailed.



\section{Experiment}

We conduct our experiments on two datasets which are PASCAL3D+ \cite{pascal} and ObjectNet3D \cite{ObjectNet3D}. 
The experiments are implemented under the general semi-supervised setting and the experimental results are compared with the supervised rotation regression methods ~\cite{matFisher} and \cite{rotLaplace}, as well as the semi-supervised rotation regression method FisherMatch \cite{fishermatch}.

\subsection{Datasets}\label{sec:data}

\textbf{PASCAL3D+} \cite{pascal} is a 3D object detection and rotation regression dataset provided by the Stanford CVGL lab. It extends the 12 categories of PASCAL VOC 2012 \cite{pascalVOC} by adding 3D annotations and includes additional images from ImageNet \cite{russakovsky2015imagenet}. We use the original training-test split of the dataset and further divide the training set into a labeled training set with ground truth annotations and an unlabeled training set with only image and category information. During the training and testing process, we only use the ImageNet portion of the data, excluding the PASCAL data and additional synthetic data. The ratio of labeled data is set as 5\%, 10\%, and 20\% of the training set.

\textbf{ObjectNet3D} \cite{ObjectNet3D} is a comprehensive dataset designed for advancing 3D object recognition in computer vision. It includes a diverse collection of 90,127 images spanning 100 categories, with each image meticulously annotated to align objects with corresponding 3D shapes. This alignment facilitates precise 3D pose and shape annotations, making ObjectNet3D instrumental for tasks such as 3D object rotation regression and image-based 3D shape retrieval. Following the split of ObjectNet3D, we directly use the training dataset for training and the val dataset for validation. The ratio of labeled training data is set as 5\%, 10\%, 20\% of the training set.

\subsection{Implementation Details}\label{sec:detail}

\textbf{Baseline}. We compare the results with two supervised methods: \cite{matFisher}(Sup.-Fisher) and \cite{rotLaplace}(SSL-rot.Laplace) and one semi-supervised method \cite{LaplaceMatch}(Sup.-Fisher). We reproduce the baselines in SSL setting using the best parameters provided on the GitHub pages of these models to train and test the networks with labeled data proportions of 5\%, 10\%, and 20\% of the total training samples.

\textbf{Experimental Details}. The backbone is ResNet18 \cite{he2015deep}. The learning rate is \( 10^{-4} \). Moreover, in the semi-supervised learning phase, the learning rate is reduced to \( 10^{-5} \) for PSCAL3D+. We set a maximum of 20k training iterations, and 10k of them are for supervised training. For the supervised stage, we train the model with a batchsize of 32, and for the SSL stage, there are 32 labeled samples and 128 unlabeled samples for each training batch. The entropy threshold is \( \tau = -3.9 \). For the experiments with multi-stage curriculum learning, the $\alpha_\text{start}(\%)$ and $\alpha_\text{end}(\%)$ is set as 65 and 95 respectively, and the $n_\text{stage}$ is 4. For the experiments with adaptive curriculum learning, $\tau_\text{start}$ is -4.5 and $\tau_\text{end}$ is -3.9. The patches number $n$ for {\augName} is set as 5, and the augmentation pool $\{\mathcal{A}\}$ is a selected one of the methods with ACC@30$^\circ$ > $76\%$ in Figure \ref{fig:aug_ablation}.

To address the samples of different categories at one time, there is an embedding branch in our model to handle multi-category data. The embedding layer maps the image category ID to a vector of \( N_{embedding} \) dimensions, which is then concatenated with the features obtained from ResNet18 and passed into the final rotation regression prediction head. The \( N_{embedding} \) for PASCAL3D+ data is 32, and for ObjectNet3D it is 256. 

\begin{table*}[!htbp]
    \centering
    \caption{The average 3D rotation regression performance (\textbf{Mean Med($^{\circ}$)$\downarrow$/ACC@30$^{\circ}(\%)$$\uparrow$}) across all 100 categories on \textbf{ObjectNet3D} \cite{ObjectNet3D} under different ratios of labeled training data. Each value is reported as the mean $\pm$ standard deviation, computed over three independent runs. The best results are marked as \textbf{bold}.}
    \begin{tabular}{l|cccc}
    \toprule
        \multirow{2}{*}{Method} & \multicolumn{4}{c}{Ratio of Labled Samples} \\ 
        \cline{2-5}
        ~ & 5\% & 10\% & 20\% & All \\  
        \midrule
        Sup.-Fisher & 46.79{\scriptsize $\pm$0.67}/42.93{\scriptsize $\pm$0.63} & 44.85{\scriptsize $\pm$0.47}/44.49{\scriptsize $\pm$0.71} & 40.68{\scriptsize $\pm$0.63}/48.12{\scriptsize $\pm$0.38} & 27.09{\scriptsize $\pm$0.52}/59.34{\scriptsize $\pm$0.65} \\ 
        Sup.-Laplace & 45.97{\scriptsize $\pm$0.45}/44.22{\scriptsize $\pm$0.51} & 44.38{\scriptsize $\pm$0.53}/46.16{\scriptsize $\pm$0.42} & 40.18{\scriptsize $\pm$0.62}/49.71{\scriptsize $\pm$0.39} & 25.76{\scriptsize $\pm$0.57}/61.67{\scriptsize $\pm$0.73} \\   
        FisherMatch & 41.19{\scriptsize $\pm$0.54}/48.02{\scriptsize $\pm$0.78} & 36.18{\scriptsize $\pm$0.57}/52.59{\scriptsize $\pm$0.62} & 33.17{\scriptsize $\pm$0.67}/55.43{\scriptsize $\pm$0.68} & --/-- \\ 
        Ours(multi-stage) & 38.14{\scriptsize $\pm$0.64}/51.79{\scriptsize $\pm$0.47} & 35.17{\scriptsize $\pm$0.52}/53.38{\scriptsize $\pm$0.77} & 32.01{\scriptsize $\pm$0.42}/55.82{\scriptsize $\pm$0.65} & --/-- \\ 
        Ours(adaptive) & \textbf{37.26}{\scriptsize $\pm$0.51}/\textbf{52.45}{\scriptsize $\pm$0.58} & \textbf{34.89}{\scriptsize $\pm$0.53}/\textbf{53.93}{\scriptsize $\pm$0.66} & \textbf{31.58}{\scriptsize $\pm$0.69}/\textbf{55.97}{\scriptsize $\pm$0.59} & --/-- \\
    \bottomrule
    \end{tabular}
    \label{tab:res-obj}
\end{table*}

\textbf{Evaluation Metrics}. Two general metrics are used in our experiments to evaluate the error between model predictions and ground truth. 
The first one is median which refers to the median of the angular prediction errors for all samples within a category. 
The second one is ACC@30$^{\circ}$, and it considers samples with a prediction error within 30 degrees to be correctly predicted. ACC@30$^{\circ}$ for a category is calculated by dividing the number of correctly predicted samples by the total number of samples in the test dataset for that category.

For models trained on all categories of a dataset together, following \cite{fishermatch}, we use the mean median (Mean Med) and mean ACC@30$^{\circ}$ (ACC@30$^{\circ}$) to evaluate the performance for the angular estimation of multiple categories. Mean Med and ACC@30$^{\circ}$ are the average values of the median error and the proportion of test samples with prediction errors less than 30 degrees across multiple categories, respectively.

Using \( R \) to represent the model's predicted rotation matrix and \( R_{\text{gt}} \) to represent the ground truth label, the geodesic distance can be expressed as:
\begin{equation}
    d\left(R_{gt},R\right)=\arccos\left(\frac{tr\left(R_{gt}{}^{\top}R\right)-1}{2}\right)/\pi,
\end{equation}
and the angle error between predicted values and ground truth (measured in the range of 0 to 180 degrees) can be expressed as:
\begin{equation}
    \begin{aligned}
err^{\circ} &= d\left(R_{gt},R\right)\times180 \\
    &=\arccos\left(\frac{tr\left(R_{gt}{}^{\top}R\right)-1}{2}\right)/\pi\times180 ,
\end{aligned}
\end{equation}
the accuracy of class $c$ is:
\begin{equation}
    Acc.=
\begin{cases}
    1, & \mathrm{~if~} d(R_{gt},R)<\lambda \mathrm{~and~} c=c_{gt}  \\
    0, & \mathrm{~otherwise}.
\end{cases}
\end{equation}

\subsection{Quantitative Results}\label{sec:results}

Tables \ref{tab:res-pascal} and \ref{tab:res-obj} present the quantitative results of our method on the PASCAL3D+ and ObjectNet3D datasets, respectively. Our proposed approaches, both \textbf{Ours(multi-stage)} and \textbf{Ours(adaptive)}, consistently outperform three baselines across all labeled data ratios (5\%, 10\%, and 20\%). 

Specifically, our methods demonstrate a improvement, particularly in low-data regimes. For instance, on PASCAL3D+ with only 5\% labeled data, our adaptive method reduces the Mean Med error from 16.48$^{\circ}$ to 14.17$^{\circ}$ and increases ACC@30$^{\circ}$ from 74.73\% to 79.24\%, a substantial gain of over 4.5 percentage points. This highlights the effectiveness of our proposed method in leveraging unlabeled data.

From an internal comparison perspective, the adaptive method often yields slightly better or comparable performance to the multi-stage approach, demonstrating its efficiency and robustness as a simpler yet powerful alternative. These results validate that incorporating hardness-aware curriculum learning and {\augName} is a highly effective strategy for semi-supervised rotation regression.

\begin{table}[!htbp]
    \centering
    \caption{The model performance with different Curriculum Pseudo Labeling strategies.}
    \begin{tabular}{ccc|cc}
    \toprule
        Aug & CL-M & CL-A & Mean Med$\downarrow$ & ACC@30$^{\circ}$$\uparrow$  \\ \midrule
        {} & {} & {} & 16.54 & 74.68  \\ 
        {} & \ding{51} & {} & 15.17 & 76.86  \\ 
        {} & {} & \ding{51} & 15.02 & 76.84  \\ 
        \ding{51} & {} & {} & 14.31 & 78.74  \\ 
        \ding{51} & \ding{51} & {} & 14.22 & 79.04  \\  
        \ding{51} & {} & \ding{51} & \textbf{14.17} & \textbf{79.24}  \\ 
        \bottomrule
    \end{tabular}
    \label{tab:ablation}
\end{table}

\subsection{Ablation Studies}\label{sec:ablation}

We conduct training on the PASCAL3D+ dataset using 5\% of the labeled data and use ResNet18 as the backbone for the ablation studies. 

\subsubsection{Effect of Different Modules}
Through this ablation study, we verify that each component of our proposed framework contributes positively to the overall performance. The results are detailed in Table \ref{tab:ablation}.

The first row shows the baseline performance without any of our proposed modules, achieving a Mean Med of 16.54$^{\circ}$ and ACC@30$^{\circ}$ of 74.68\%. 
Rows 2 and 3 demonstrate the standalone impact of our curriculum learning strategies. Introducing multi-stage curriculum learning (CL-M) or adaptive curriculum learning (CL-A) alone improves performance, reducing the Mean Med error to 15.17$^{\circ}$ and 15.02$^{\circ}$, respectively. This confirms the value of guiding the learning process from easy to hard samples.
Row 4 shows that applying our strong data augmentation strategy ({\augName}) by itself provides the most substantial single-component boost, lowering the error to 14.31$^{\circ}$ and increasing accuracy to 78.74\%. This highlights the critical role of robust data augmentation in consistency-based semi-supervised learning.
Finally, rows 5 and 6 illustrate the synergistic effect of combining strong augmentation with curriculum learning. When {\augName} is paired with either CL-M or CL-A, we observe the best performance. The combination of strong augmentation and our adaptive curriculum learning (CL-A) yields the top result, with a Mean Med of \textbf{14.17$^{\circ}$} and an ACC@30$^{\circ}$ of \textbf{79.24\%}. This ablation study clearly validates that both our advanced augmentation and curriculum learning modules are essential and complementary components of our final model.

\subsubsection{Data Augmentation}\label{sec:abl_aug}

\paragraph{\textbf{Comparison of different augmentation method.}}

This experiment aims to investigate the impact of various advanced strong data augmentation strategies on the performance of semi-supervised rotation estimation models. The methods compared include CutOut~\cite{devries2017improved}, CutMix~\cite{cutmix}, CutOcc and YOCO~\cite{yoco}.
CutOut randomly masks out small rectangular patches from the input image, thereby encouraging the model to focus on the surrounding context. CutMix, on the other hand, replaces a random region of the image with a patch from another image, and adjusts the label accordingly. CutOcc is a hybrid strategy that combines the effects of CutOut and CutMix. YOCO performs region-wise augmentation and recomposition to maintain semantic coherence while diversifying image structure. 

\begin{table}[!htbp]
    \centering
    \caption{The model's performance with different advanced strong augmentations.}
    \begin{tabular}{l|cc}
        \toprule
        Hard\_Aug & Mean Med$\downarrow$ & ACC@30$^{\circ}$$\uparrow$ \\ \midrule
        -- & 16.54 & 74.68  \\ 
        CutOut & 16.84 & 75.20  \\ 
        CutMix & 16.21 & 75.68  \\ 
        CutOcc & 16.02 & 75.47  \\ 
        YOCO & 15.31 & 76.66  \\ 
        Ours(all)  & 15.11 & 77.35  \\ 
        Ours(selected $\{\mathcal{A}\}$) & \textbf{14.31} & \textbf{78.74} \\
        \bottomrule
    \end{tabular}
    \label{tab:hard_aug}
\end{table}

\begin{figure}
    \centering
    \includegraphics[width=0.8\linewidth]{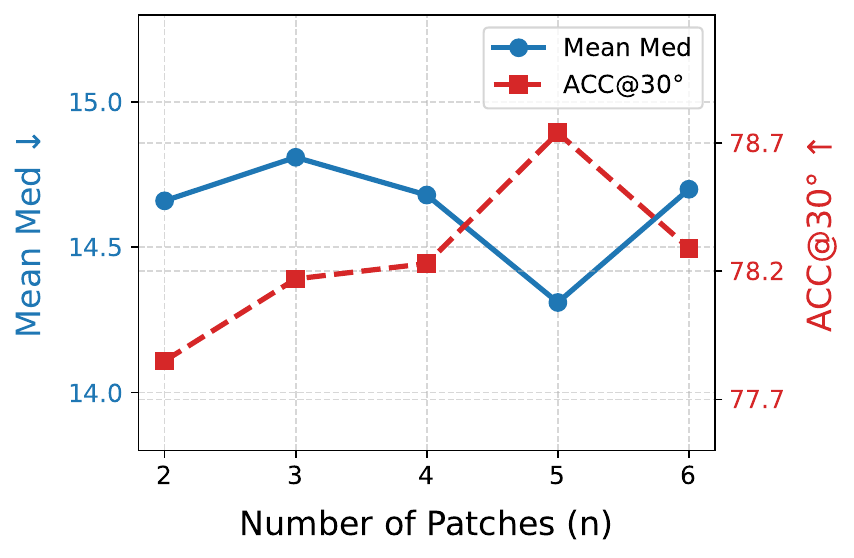}
    \caption{Effect of the number of patch splits ($n$) on model performance, with $\alpha_{\text{start}}$ fixed at 65 and $\alpha_{\text{end}}$ fixed at 95.}
    \label{fig:n_split}
\end{figure}

For our proposed method {\augName}, we evaluate two variants of its augmentation pool ($\{\mathcal{A}\}$): 
\begin{itemize}
    \item \textbf{Ours(all)}: Utilizes all 16 augmentation methods explored in Figure \ref{fig:aug_ablation}.
    \item \textbf{Ours(selected $\{\mathcal{A}\}$)}: Employs a selected pool of the 7 best-performing augmentations (those with ACC@30$^\circ$ > $76\%$ in Figure \ref{fig:aug_ablation}, namely \textit{invert}, \textit{rotate}, \textit{brightness}, \textit{shearX}, \textit{shearY}, \textit{translateX}, and \textit{translateY}).
\end{itemize}

As shown in Table~\ref{tab:hard_aug}, both of our proposed {\augName} ~variants outperform other methods. \textbf{Ours(all)} reduces the Mean Med to 15.11$^\circ$ and improves ACC@30$^\circ$ to 77.35\%, surpassing YOCO. More impressively, \textbf{Ours(selected $\mathcal{A}$)} achieves the best results among all evaluated techniques, reducing the Mean Med to \textbf{14.31$^\circ$} and improving the ACC@30$^\circ$ to \textbf{78.74\%}. This demonstrates that while a diverse augmentation pool is beneficial, a carefully selected set of high-quality, task-appropriate augmentations is even more effective. This is likely because it avoids augmentations that might introduce ambiguity or disrupt critical structural features for rotation estimation.

\paragraph{\textbf{Patches Number $n$.}}

When applying data augmentation, increasing the number of patches into which an image is divided generally introduces greater diversity during training. However, this also increases the semantic gap between sub-regions, potentially affecting the consistency and feature alignment for the model. To explore whether increasing the number of patches consistently improves semi-supervised rotation estimation, we compare the model's performance under different numbers of patch splits $n$, where each augmented image is composed of $n$ sub-regions, each transformed by a different augmentation.


\begin{figure*}[!htbp]
    \centering
    \begin{subfigure}{0.32\textwidth}
        \centering
        \includegraphics[width=\linewidth]{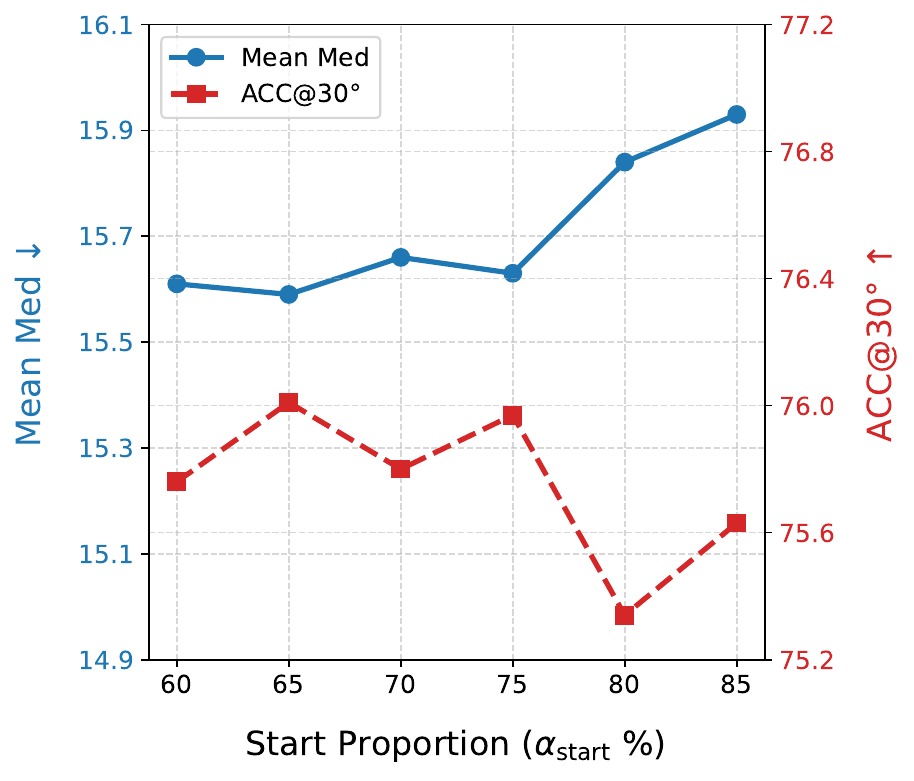}
        \caption{Effect of start proportion ($\alpha_{\text{start}}$).}
        \label{fig:ablation_c_start}
    \end{subfigure}
    \hfill
    \begin{subfigure}{0.32\textwidth}
        \centering
        \includegraphics[width=\linewidth]{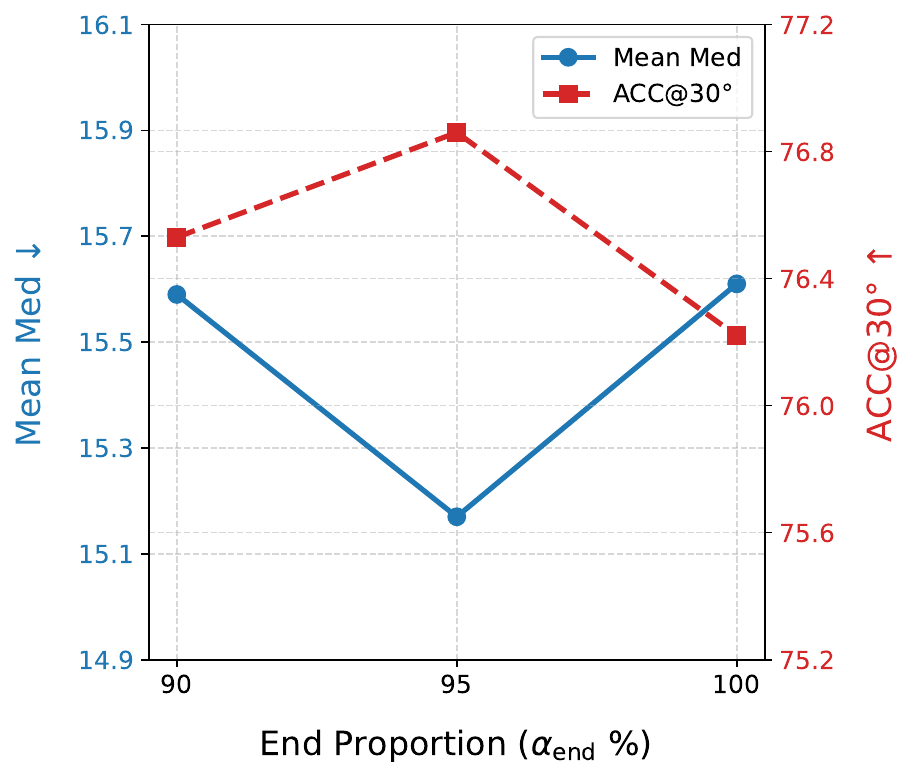}
        \caption{Effect of end proportion ($\alpha_{\text{end}}$).}
        \label{fig:ablation_c_end}
    \end{subfigure}
    \hfill
    \begin{subfigure}{0.32\textwidth}
        \centering
        \includegraphics[width=\linewidth]{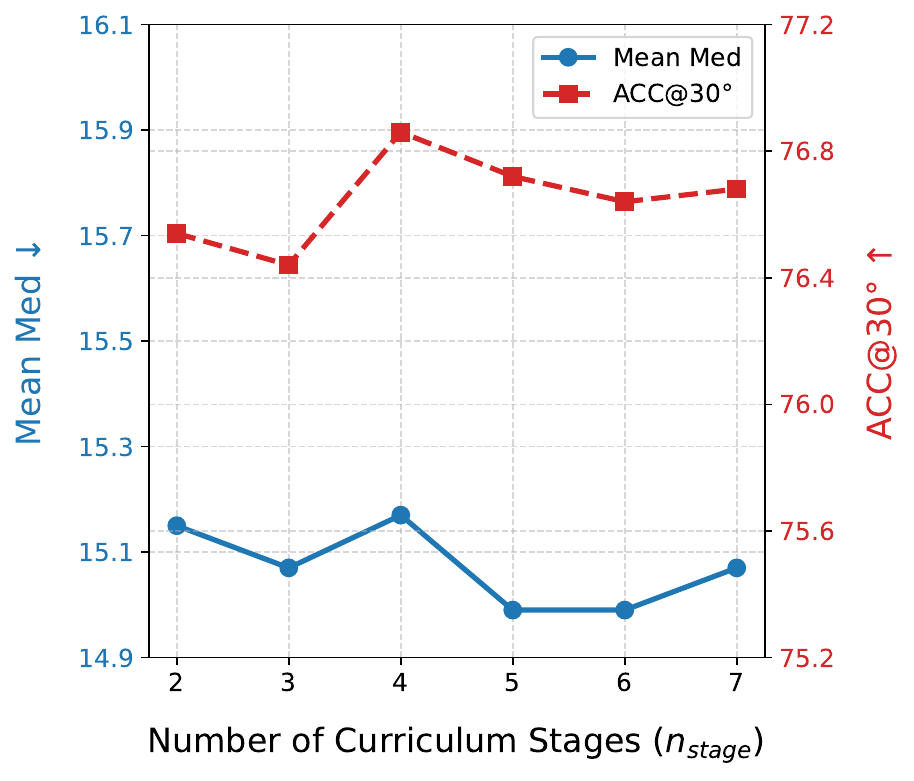}
        \caption{Effect of number of stages ($n_{\text{stage}}$).}
        \label{fig:ablation_n_stage}
    \end{subfigure}
    \caption{Ablation study on the hyperparameters of the multi-stage curriculum learning strategy. All experiments were conducted on PASCAL3D+ with 5\% labeled data, and no strong augmentation is applied. (a) Impact of the starting sample proportion ($\alpha_{\text{start}}$), with $\alpha_{\text{end}}$ fixed at 100 and $n_{\text{stage}}=4$. (b) Impact of the ending sample proportion ($\alpha_{\text{end}}$), with $\alpha_{\text{start}}$ fixed at 65 and $n_{\text{stage}}=4$. (c) Impact of the number of curriculum stages ($n_{\text{stage}}$), with $\alpha_{\text{start}}=65$, $\alpha_{\text{end}}=95$, and no strong augmentation.}
    \label{fig:ablation_multistage_cl}
\end{figure*}

\begin{figure}
    \centering
    \begin{subfigure}[b]{0.45\textwidth}
        \includegraphics[width=\linewidth]{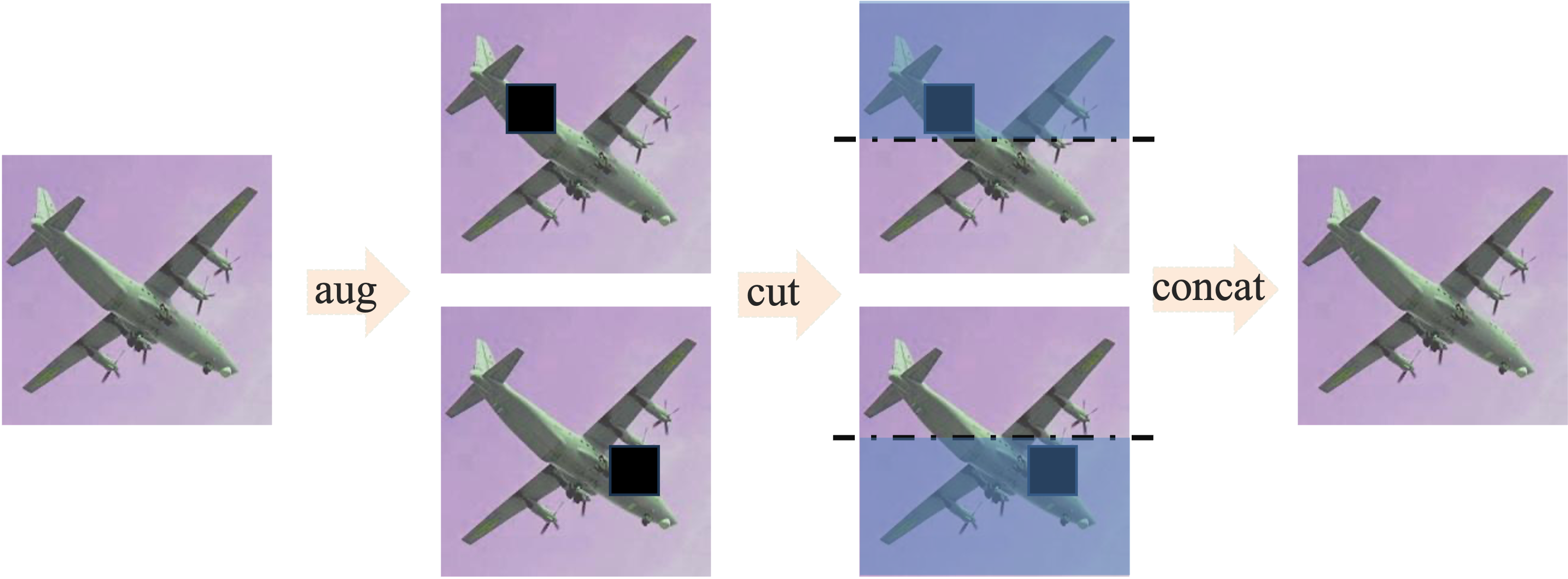}
        \caption{Cut}
        \label{fig:val_cut}
    \end{subfigure} 

    \vspace{0.5em}
    \begin{subfigure}[b]{0.45\textwidth}
        \includegraphics[width=\linewidth]{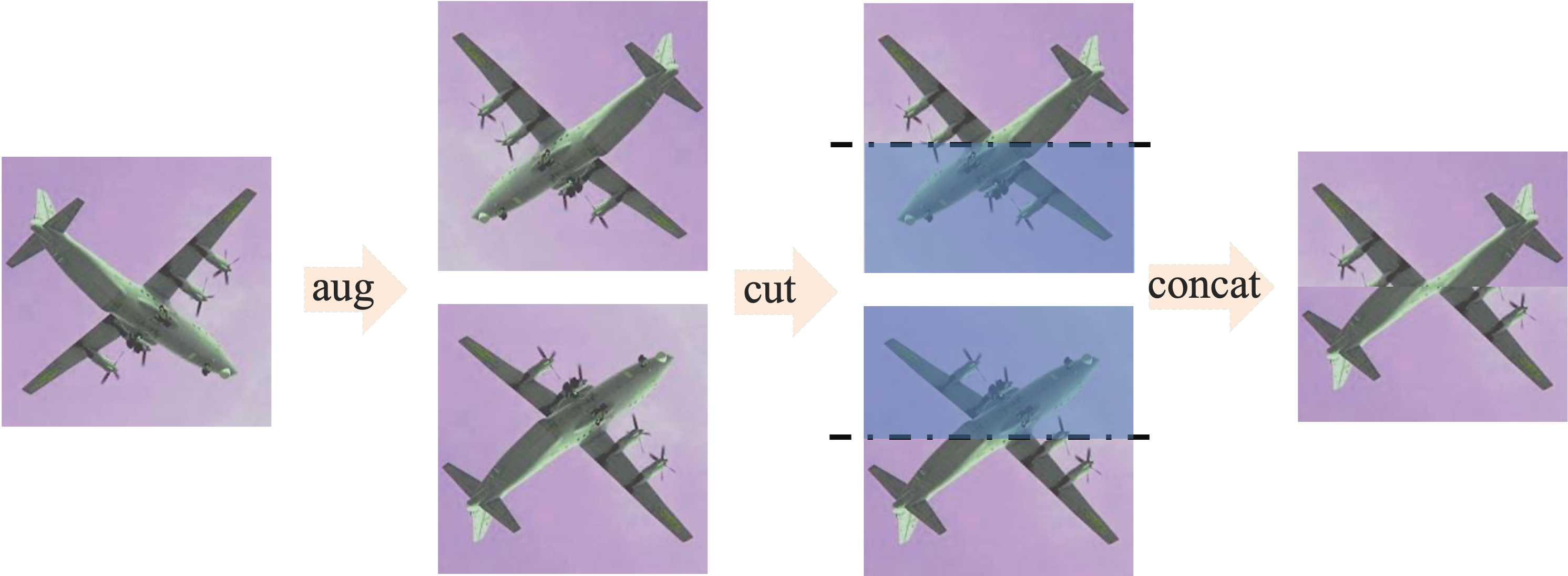}
        \caption{Flip}
        \label{fig:val_flip}
    \end{subfigure}
    \caption{Illustration of incompatible augmentation methods in \augName.}
\end{figure}

The results are summarized in Figure~\ref{fig:n_split}, increasing the number of patches generally improves performance in both Mean Med and ACC@30$^{\circ}$. Notably, the best result is achieved when $n=5$, reaching a Mean Med of 14.31$^{\circ}$ and ACC@30$^{\circ}$ of 78.74\%. This suggests that using a moderate number of patches provides a good balance between data diversity and semantic consistency. However, further increasing $n$ beyond this point does not lead to non-trivial gains, indicating a potential saturation or even semantic dilution of informative features.

These results validate the effectiveness of our patch-based augmentation strategy and provide guidance for choosing the optimal patch number in practice.

\begin{table}[!htbp]
    \centering
    \resizebox{\linewidth}{!}{
    \begin{tabular}{l|cccccc}
    \toprule
        \textbf{Numbers of Patches} & 1 & 2 & 3 & 4 & 5 & 6 \\
        \textbf{Training Time(min)} & 394 & 401 & 409 & 418 & 434 & 440 \\
    \bottomrule
    \end{tabular}}
    \caption{Training time (in minutes) under different numbers of patches when applying \augName.}
    \label{tab:aug_val}
\end{table}

\paragraph{\textbf{Training Efficiency.}} 
{To evaluate the computational overhead introduced by \augName, we record the total training time under different numbers of patches ($n$), as presented in Table \ref{tab:aug_val}. The baseline model without spatial patch assembly ($n=1$) takes approximately 394 minutes to train. As the number of patches increases, the training time only grows marginally. For instance, under our optimal setting ($n=5$), the total training time is 434 minutes, which introduces merely a $\sim$10.15\% increase compared to the baseline. }

{This high efficiency stems from the design of \augName. The image cropping and concatenation operations are lightweight and executed entirely on the CPU during the data loading phase. Furthermore, since the final composite image strictly preserves the original input dimensions ($W \times H \times C$), it introduces zero additional computational burden or memory footprint to the GPU during the network's forward and backward passes. Therefore, \augName is a highly practical augmentation strategy, offering substantial performance gains at a negligible and highly acceptable computational cost.}

\paragraph{\textbf{Validity of Specific Augmentations.}}
{While \augName successfully enhances feature diversity for most image augmentation types, it can be ineffective or even detrimental when applied to certain transformations. Specifically, the patch-level cutting and concatenating mechanism introduces two potential issues. First, spatially localized augmentations may be inadvertently discarded if the modified region falls outside the selected patch. Second, augmentations that alter the global spatial orientation of the object can lead to severe structural incoherence when stitched together. Figures \ref{fig:val_cut} and \ref{fig:val_flip} visually illustrate these two failure modes.}

\begin{itemize}
    \item \textbf{Cutout (Ineffective):} {As shown in Figure \ref{fig:val_cut}, the final composite aeroplane is identical to the original image. This occurs because the `mosaic' operation may discard the specific region modified by the augmentation (the black `cutout' area), effectively nullifying the transformation and wasting computational resources. }
    
    \item \textbf{Flip (Detrimental):} {Figure \ref{fig:val_flip} shows an anomalous aeroplane with two tails and no head. This issue arises when independent random flips result in images with opposing orientations. The subsequent `cut and concatenate' process combines them into a structurally incoherent object, destroying the holistic geometric cues and making orientation regression impossible.} 
\end{itemize}

\subsubsection{Multi-stage Curriculum Pseudo Labeling}


\begin{figure}
    \centering
    \includegraphics[width=0.8\linewidth]{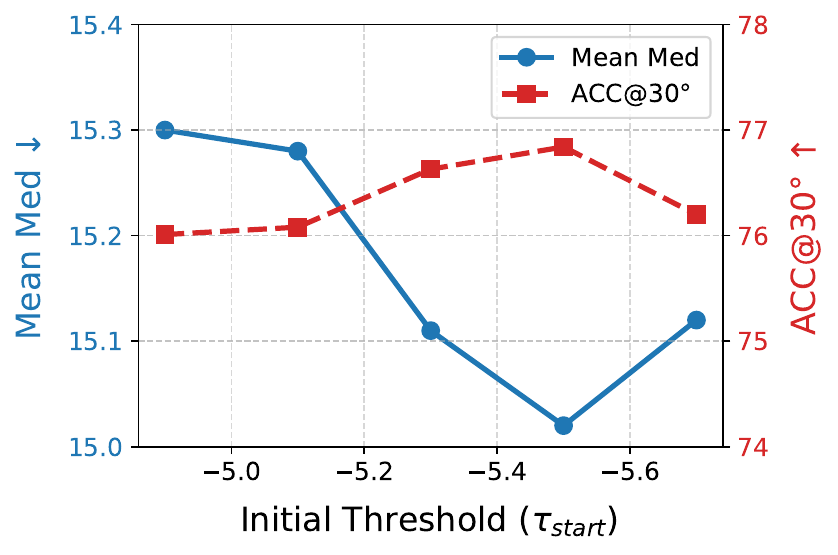}
    \caption{Effect of the initial threshold ($\tau_\text{start}$) on performance in our adaptive CL strategy.}
    \label{fig:table8}
\end{figure}

\paragraph{\textbf{Start proportion.}} 
We test how different starting proportions \( \alpha_{\text{start}} \) affect the model's performance, with \( \alpha_{\text{end}} \) fixed at 100. As shown in Figure \ref{fig:ablation_multistage_cl}(a), the choice of \( \alpha_{\text{start}} \) has a noticeable impact on the final performance. The results show that performance is strongest when \( \alpha_{\text{start}} = 65 \), achieving both the lowest Mean Med error and the highest ACC@30$^{\circ}$. This suggests that starting the curriculum with the top 65\% of `easy' samples provides an optimal balance, effectively guiding the model without exposing it to the most challenging samples too early in the training process.

\paragraph{\textbf{End proportion.}} After identifying \(\alpha_{\text{start}}=65\) as optimal, we found that combining CPL with our advanced strong augmentation required further tuning. We fix \( \alpha_{\text{start}} \) at 65 and examine the impact of different \(\alpha_{\text{end}}\) values. The experimental results in Figure \ref{fig:ablation_multistage_cl}(b) indicate that setting \(\alpha_{\text{end}} = 95\) achieves better results than \(\alpha_{\text{end}} = 100\). This implies that excluding the most difficult 5\% of samples throughout the entire training process can prevent the model from learning from potentially noisy or unreliable pseudo-labels, ultimately boosting performance.

\paragraph{\textbf{Stage numbers.}}

To evaluate the effect of the number of training stages in fixed curriculum learning, we conducted experiments by fixing the curriculum start point $\alpha_{\text{start}} = 65$ and end point $\alpha_{\text{end}} = 95$, and varying the number of learning stages $n_{\text{stage}}$. Note that in this experiment, no strong data augmentation was applied during training to isolate the impact of stage division.

As shown in Figure~\ref{fig:ablation_multistage_cl}(c), varying the number of curriculum learning stages yields minor fluctuations in both Mean Med and ACC@30$^\circ$. The best accuracy is achieved when $n_{\text{stage}} = 4$, while the lowest Mean Med is obtained when $n_{\text{stage}} = 5$ and $6$. These results indicate that increasing the stage resolution beyond a certain point does not consistently improve performance and may lead to marginal overfitting or redundancy. Thus, we conclude that using 4 to 5 stages provides a good trade-off between training complexity and model effectiveness.

\subsubsection{Adaptive Curriculum Learning}

Figure~\ref{fig:table8} compares the effects of different initial thresholds $\tau_\text{start}$ on the performance of semi-supervised rotation estimation when using our adaptive threshold scheduling strategy. In this experiment, we fixed the final threshold $\tau_\text{end} = -3.9$ to keep consistent with the baseline and did not apply the {\augName} method to unlabeled samples. We evaluate a range of $\tau_\text{start}$ values from $-4.9$ to $-5.7$.

As shown in Figure~\ref{fig:table8}, different values of $\tau_\text{start}$ significantly influence model performance. When $\tau_\text{start} = -5.5$, the model achieves the best results, with the lowest Mean Med (15.02$^{\circ}$) and the highest ACC@30$^\circ$ (76.84\%). This suggests that setting a relatively lower initial threshold allows the model to focus more on high-entropy pseudo labels in the early stages, which benefits training stability and accuracy. However, setting $\tau_\text{start}$ too low (e.g., $-5.7$) does not further improve performance, indicating a saturation effect in threshold sensitivity.

\section{Discussion and Conclusion}

In this paper, we mainly address the problem of rigid pseudo-label filtering in semi-supervised rotation regression, which largely limits the effective utilization of unlabeled data. Our solution introduces a hardness-aware curriculum learning framework termed HACMatch that dynamically selects samples based on difficulty with two different strategies. We further propose a structured data augmentation technique termed {\augName} that aims at preserving geometric integrity while providing necessary feature diversity for rotation estimation. Extensive experimental results on PASCAL3D+ and ObjectNet3D datasets demonstrate improvements over existing methods, particularly in low-data scenarios, validating our approach's effectiveness in leveraging unlabeled data for accurate rotation regression. 

Looking ahead, future work could focus on developing curriculum strategies that respond directly to the model's learning state, reducing the need for pre-defined parameters. Furthermore, automatically optimizing the augmentation pool for {\augName} presents a promising direction to enhance its generalization and effectiveness across diverse tasks. 

{Finally, exploring the potential of Multi-modal Large Language Models (MLLMs) and Large Language Models (LLMs) presents an exciting frontier for this task. Recently, LLMs and MLLMs have shown remarkable progress in providing robust semantic priors and reasoning capabilities for semi-supervised learning across various domains, such as sentiment analysis~\cite{li2024semantic}, 3D medical image segmentation~\cite{kumari2024leveraging}, and explainable video anomaly detection~\cite{mumcu2025leveraging}. For semi-supervised 3D rotation regression, MLLMs could potentially provide rich, task-specific geometric descriptions or contextual object priors to assist in aligning and rectifying pseudo-labels. This would be particularly beneficial for challenging, heavily occluded, or out-of-distribution instances, paving the way for more robust and intelligent 3D vision systems.}


\printcredits

\section*{Acknowledgments}
This paper is supported by the Fundamental Research Funds for the Central Universities (project number YG2024ZD06) and NSFC
(62176155).

\bibliographystyle{cas-model2-names}

\bibliography{references}

@String(AAAI = {AAAI})

@inproceedings{pascal,
  title={Beyond pascal: A benchmark for 3d object detection in the wild},
  author={Xiang, Yu and Mottaghi, Roozbeh and Savarese, Silvio},
  booktitle={IEEE Winter Conference on Applications of Computer Vision},
  pages={75--82},
  year={2014},
  organization={IEEE}
}

@inproceedings{ObjectNet3D,
  title={Objectnet3d: A large scale database for 3d object recognition},
  author={Xiang, Yu and Kim, Wonhui and Chen, Wei and Ji, Jingwei and Choy, Christopher and Su, Hao and Mottaghi, Roozbeh and Guibas, Leonidas and Savarese, Silvio},
  booktitle={Computer Vision--ECCV 2016: 14th European Conference, Amsterdam, The Netherlands, October 11-14, 2016, Proceedings, Part VIII 14},
  pages={160--176},
  year={2016},
  organization={Springer}
}

@article{matFisher,
  title={Probabilistic orientation estimation with matrix fisher distributions},
  author={Mohlin, David and Sullivan, Josephine and Bianchi, G{\'e}rald},
  journal={Advances in Neural Information Processing Systems},
  volume={33},
  pages={4884--4893},
  year={2020}
}

@article{rotLaplace,
  title={A Laplace-inspired Distribution on SO (3) for Probabilistic Rotation Estimation},
  author={Yin, Yingda and Wang, Yang and Wang, He and Chen, Baoquan},
  journal={The Eleventh International Conference on Learning Representations},
  year={2023}
}

@inproceedings{fishermatch,
  title={Fishermatch: Semi-supervised rotation regression via entropy-based filtering},
  author={Yin, Yingda and Cai, Yingcheng and Wang, He and Chen, Baoquan},
  booktitle={Proceedings of the IEEE Conference on Computer Vision and Pattern Recognition},
  pages={11164--11173},
  year={2022}
}

@article{LaplaceMatch,
  title={Towards Robust Probabilistic Modeling on SO (3) via Rotation Laplace Distribution},
  author={Yin, Yingda and Lyu, Jiangran and Wang, Yang and Wang, He and Chen, Baoquan},
  journal={arXiv preprint arXiv:2305.10465},
  year={2023}
}

@inproceedings{CLSemi,
  title={Curriculum labeling: Revisiting pseudo-labeling for semi-supervised learning},
  author={Cascante-Bonilla, Paola and Tan, Fuwen and Qi, Yanjun and Ordonez, Vicente},
  booktitle={Proceedings of the AAAI conference on artificial intelligence},
  volume={35},
  number={8},
  pages={6912--6920},
  year={2021}
}

@article{zhang2022flexmatch,
  title={Flexmatch: Boosting semi-supervised learning with curriculum pseudo labeling},
  author={Zhang, Bowen and Wang, Yidong and Hou, Wenxin and Wu, Hao and Wang, Jindong and Okumura, Manabu and Shinozaki, Takahiro},
  journal={Advances in Neural Information Processing Systems},
  volume={34},
  pages={18408--18419},
  year={2021}
}

@article{DeepSemiSuvey,
  title={A survey on deep semi-supervised learning},
  author={Yang, Xiangli and Song, Zixing and King, Irwin and Xu, Zenglin},
  journal={IEEE Transactions on Knowledge and Data Engineering},
  volume={35},
  number={9},
  pages={8934--8954},
  year={2022},
  publisher={IEEE}
}

@inproceedings{liu2023gen6d,
  title={Gen6d: Generalizable model-free 6-dof object pose estimation from rgb images},
  author={Liu, Yuan and Wen, Yilin and Peng, Sida and Lin, Cheng and Long, Xiaoxiao and Komura, Taku and Wang, Wenping},
  booktitle={European Conference on Computer Vision},
  pages={298--315},
  year={2022},
  organization={Springer}
}

@article{he2023onepose,
  title={Onepose++: Keypoint-free one-shot object pose estimation without CAD models},
  author={He, Xingyi and Sun, Jiaming and Wang, Yuang and Huang, Di and Bao, Hujun and Zhou, Xiaowei},
  journal={Advances in Neural Information Processing Systems},
  volume={35},
  pages={35103--35115},
  year={2022}
}

@inproceedings{wen2024foundationpose,
  title={Foundationpose: Unified 6d pose estimation and tracking of novel objects},
  author={Wen, Bowen and Yang, Wei and Kautz, Jan and Birchfield, Stan},
  booktitle={Proceedings of the IEEE/CVF Conference on Computer Vision and Pattern Recognition},
  pages={17868--17879},
  year={2024}
}

@article{auto1,
  title={Deep learning for 6D pose estimation of objects—A case study for autonomous driving},
  author={Hoque, Sabera and Xu, Shuxiang and Maiti, Ananda and Wei, Yuchen and Arafat, Md Yasir},
  journal={Expert Systems with Applications},
  volume={223},
  pages={119838},
  year={2023},
  publisher={Elsevier}
}

@inproceedings{auto2,
  title={PanelPose: A 6D Pose Estimation of Highly-Variable Panel Object for Robotic Robust Cockpit Panel Inspection},
  author={Sun, Han and Ni, Peiyuan and Li, Zhiqi and Wang, Yizhao and Zhu, Xiaoxiao and Cao, Qixin},
  booktitle={2023 IEEE/RSJ International Conference on Intelligent Robots and Systems (IROS)},
  pages={3214--3221},
  year={2023},
  organization={IEEE}
}

@article{10043016,
  title={Robotic Continuous Grasping System by Shape Transformer-Guided Multiobject Category-Level 6-D Pose Estimation},
  author={Liu, Jian and Sun, Wei and Liu, Chongpei and Zhang, Xing and Fu, Qiang},
  journal={IEEE Transactions on Industrial Informatics},
  volume={19},
  number={11},
  pages={11171--11181},
  year={2023},
  publisher={IEEE}
}

@article{mohlin2020probabilistic,
  title={Probabilistic orientation estimation with matrix fisher distributions},
  author={Mohlin, David and Sullivan, Josephine and Bianchi, G{\'e}rald},
  journal={Advances in Neural Information Processing Systems},
  volume={33},
  pages={4884--4893},
  year={2020}
}

@inproceedings{rad2018bb8,
  title={Bb8: A scalable, accurate, robust to partial occlusion method for predicting the 3d poses of challenging objects without using depth},
  author={Rad, Mahdi and Lepetit, Vincent},
  booktitle={Proceedings of the IEEE international conference on computer vision},
  pages={3828--3836},
  year={2017}
}

@inproceedings{tekin2018realtime,
  title={Real-time seamless single shot 6d object pose prediction},
  author={Tekin, Bugra and Sinha, Sudipta N and Fua, Pascal},
  booktitle={Proceedings of the IEEE conference on computer vision and pattern recognition},
  pages={292--301},
  year={2018}
}

@inproceedings{pavlakos20176dof,
  title={6-dof object pose from semantic keypoints},
  author={Pavlakos, Georgios and Zhou, Xiaowei and Chan, Aaron and Derpanis, Konstantinos G and Daniilidis, Kostas},
  booktitle={2017 IEEE international conference on robotics and automation (ICRA)},
  pages={2011--2018},
  year={2017},
  organization={IEEE}
}

@inproceedings{9636212,
  title={Category-level 6d object pose estimation via cascaded relation and recurrent reconstruction networks},
  author={Wang, Jiaze and Chen, Kai and Dou, Qi},
  booktitle={2021 IEEE/RSJ International Conference on Intelligent Robots and Systems (IROS)},
  pages={4807--4814},
  year={2021},
  organization={IEEE}
}

@article{9933183,
  title={6d-vit: Category-level 6d object pose estimation via transformer-based instance representation learning},
  author={Zou, Lu and Huang, Zhangjin and Gu, Naijie and Wang, Guoping},
  journal={IEEE Transactions on Image Processing},
  volume={31},
  pages={6907--6921},
  year={2022},
  publisher={IEEE}
}

@inproceedings{pitteri2019cornet,
  title={CorNet: generic 3D corners for 6D pose estimation of new objects without retraining},
  author={Pitteri, Giorgia and Ilic, Slobodan and Lepetit, Vincent},
  booktitle={Proceedings of the IEEE/CVF International Conference on Computer Vision Workshops},
  pages={0--0},
  year={2019}
}

@article{gou2022unseen,
  title={Unseen object 6D pose estimation: a benchmark and baselines},
  author={Gou, Minghao and Pan, Haolin and Fang, Hao-Shu and Liu, Ziyuan and Lu, Cewu and Tan, Ping},
  journal={arXiv preprint arXiv:2206.11808},
  year={2022}
}

@inproceedings{murphy2022implicitpdf,
  title={Implicit-PDF: Non-Parametric Representation of Probability Distributions on the Rotation Manifold},
  author={Murphy, Kieran A and Esteves, Carlos and Jampani, Varun and Ramalingam, Srikumar and Makadia, Ameesh},
  booktitle={International Conference on Machine Learning},
  pages={7882--7893},
  year={2021},
  organization={PMLR}
}

@inproceedings{liu2023delving,
  title={Delving into discrete normalizing flows on so (3) manifold for probabilistic rotation modeling},
  author={Liu, Yulin and Liu, Haoran and Yin, Yingda and Wang, Yang and Chen, Baoquan and Wang, He},
  booktitle={Proceedings of the IEEE/CVF Conference on Computer Vision and Pattern Recognition},
  pages={21264--21273},
  year={2023}
}

@article{klee2023image,
  title={Image to Sphere: Learning Equivariant Features for Efficient Pose Prediction},
  author={Klee, David M and Biza, Ondrej and Platt, Robert and Walters, Robin},
  journal={International Conference on Learning Representations},
  year={2023},
  organization={International Conference on Learning Representations}
}

@inproceedings{howell2023equivariant,
  title={Equivariant single view pose prediction via induced and restricted representations},
  author={Howell, Owen and Klee, David and Biza, Ondrej and Zhao, Linfeng and Walters, Robin},
  booktitle={Proceedings of the 37th International Conference on Neural Information Processing Systems},
  pages={47251--47263},
  year={2023}
}

@article{xiao2022fewshot,
  title={Few-shot object detection and viewpoint estimation for objects in the wild},
  author={Xiao, Yang and Lepetit, Vincent and Marlet, Renaud},
  journal={IEEE transactions on pattern analysis and machine intelligence},
  volume={45},
  number={3},
  pages={3090--3106},
  year={2022},
  publisher={IEEE}
}

@article{fu2022categorylevel,
  title={Category-level 6d object pose estimation in the wild: A semi-supervised learning approach and a new dataset},
  author={Fu, Yang and Wang, Xiaolong},
  journal={Advances in Neural Information Processing Systems},
  volume={35},
  pages={27469--27483},
  year={2022}
}

@inproceedings{xiao2021posecontrast,
  title={PoseContrast: Class-agnostic object viewpoint estimation in the wild with pose-aware contrastive learning},
  author={Xiao, Yang and Du, Yuming and Marlet, Renaud},
  booktitle={2021 International Conference on 3D Vision (3DV)},
  pages={74--84},
  year={2021},
  organization={IEEE}
}

@article{LIU2024110151,
  title={PA-Pose: Partial point cloud fusion based on reliable alignment for 6D pose tracking},
  author={Liu, Zhenyu and Wang, Qide and Liu, Daxin and Tan, Jianrong},
  journal={Pattern Recognition},
  volume={148},
  pages={110151},
  year={2024},
  publisher={Elsevier}
}

@article{FixMatch,
  title={Fixmatch: Simplifying semi-supervised learning with consistency and confidence},
  author={Sohn, Kihyuk and Berthelot, David and Carlini, Nicholas and Zhang, Zizhao and Zhang, Han and Raffel, Colin A and Cubuk, Ekin Dogus and Kurakin, Alexey and Li, Chun-Liang},
  journal={Advances in neural information processing systems},
  volume={33},
  pages={596--608},
  year={2020}
}

@article{odena2016semisupervised,
  title={Semi-supervised learning with generative adversarial networks},
  author={Odena, Augustus},
  journal={arXiv preprint arXiv:1606.01583},
  year={2016}
}

@article{radford2016unsupervised,
  title={Unsupervised representation learning with deep convolutional generative adversarial networks},
  author={Radford, Alec and Metz, Luke and Chintala, Soumith},
  journal={arXiv preprint arXiv:1511.06434},
  year={2015}
}

@article{goodfellow2014generative,
  title={Generative adversarial networks},
  author={Goodfellow, Ian and Pouget-Abadie, Jean and Mirza, Mehdi and Xu, Bing and Warde-Farley, David and Ozair, Sherjil and Courville, Aaron and Bengio, Yoshua},
  journal={Communications of the ACM},
  volume={63},
  number={11},
  pages={139--144},
  year={2020},
  publisher={ACM New York, NY, USA}
}

@article{kingma2022autoencoding,
  title={Auto-encoding variational bayes},
  author={Kingma, Diederik P and Welling, Max},
  journal={arXiv preprint arXiv:1312.6114},
  year={2013}
}

@inproceedings{park2017adversarial,
  title={Adversarial dropout for supervised and semi-supervised learning},
  author={Park, Sungrae and Park, JunKeon and Shin, Su-Jin and Moon, Il-Chul},
  booktitle={Proceedings of the AAAI conference on artificial intelligence},
  volume={32},
  number={1},
  year={2018}
}

@article{xie2020unsupervised,
  title={Unsupervised data augmentation for consistency training},
  author={Xie, Qizhe and Dai, Zihang and Hovy, Eduard and Luong, Thang and Le, Quoc},
  journal={Advances in neural information processing systems},
  volume={33},
  pages={6256--6268},
  year={2020}
}

@inproceedings{iscen2019label,
  title={Label propagation for deep semi-supervised learning},
  author={Iscen, Ahmet and Tolias, Giorgos and Avrithis, Yannis and Chum, Ondrej},
  booktitle={Proceedings of the IEEE/CVF conference on computer vision and pattern recognition},
  pages={5070--5079},
  year={2019}
}

@inproceedings{9157594,
  title={Data-efficient semi-supervised learning by reliable edge mining},
  author={Chen, Peibin and Ma, Tao and Qin, Xu and Xu, Weidi and Zhou, Shuchang},
  booktitle={Proceedings of the IEEE/CVF Conference on Computer Vision and Pattern Recognition},
  pages={9192--9201},
  year={2020}
}

@inproceedings{li2020densityaware,
  title={Density-aware graph for deep semi-supervised visual recognition},
  author={Li, Suichan and Liu, Bin and Chen, Dongdong and Chu, Qi and Yuan, Lu and Yu, Nenghai},
  booktitle={Proceedings of the IEEE/CVF Conference on Computer Vision and Pattern Recognition},
  pages={13400--13409},
  year={2020}
}

@inproceedings{trinet,
  title={Tri-net for semi-supervised deep learning},
  author={Dong-DongChen, W and WeiGao, ZH},
  booktitle={Proceedings of twenty-seventh international joint conference on artificial intelligence},
  pages={2014--2020},
  year={2018}
}

@incollection{bengio2012evolving,
  title={Evolving culture versus local minima},
  author={Bengio, Yoshua},
  booktitle={Growing adaptive machines: Combining development and learning in artificial neural networks},
  pages={109--138},
  year={2014},
  publisher={Springer}
}

@article{wang2021survey,
  title={A survey on curriculum learning},
  author={Wang, Xin and Chen, Yudong and Zhu, Wenwu},
  journal={IEEE transactions on pattern analysis and machine intelligence},
  volume={44},
  number={9},
  pages={4555--4576},
  year={2021},
  publisher={IEEE}
}

@inproceedings{yoco,
  title={You only cut once: Boosting data augmentation with a single cut},
  author={Han, Junlin and Fang, Pengfei and Li, Weihao and Hong, Jie and Armin, Mohammad Ali and Reid, Ian and Petersson, Lars and Li, Hongdong},
  booktitle={International Conference on Machine Learning},
  pages={8196--8212},
  year={2022},
  organization={PMLR}
}

@article{russakovsky2015imagenet,
  title={Imagenet large scale visual recognition challenge},
  author={Russakovsky, Olga and Deng, Jia and Su, Hao and Krause, Jonathan and Satheesh, Sanjeev and Ma, Sean and Huang, Zhiheng and Karpathy, Andrej and Khosla, Aditya and Bernstein, Michael and others},
  journal={International journal of computer vision},
  volume={115},
  pages={211--252},
  year={2015},
  publisher={Springer}
}

@inproceedings{he2015deep,
  title={Deep residual learning for image recognition},
  author={He, Kaiming and Zhang, Xiangyu and Ren, Shaoqing and Sun, Jian},
  booktitle={Proceedings of the IEEE conference on computer vision and pattern recognition},
  pages={770--778},
  year={2016}
}

@article{devries2017improved,
  title={Improved regularization of convolutional neural networks with cutout},
  author={DeVries, Terrance and Taylor, Graham W},
  journal={arXiv preprint arXiv:1708.04552},
  year={2017}
}

@inproceedings{cutmix,
  title={Cutmix: Regularization strategy to train strong classifiers with localizable features},
  author={Yun, Sangdoo and Han, Dongyoon and Oh, Seong Joon and Chun, Sanghyuk and Choe, Junsuk and Yoo, Youngjoon},
  booktitle={Proceedings of the IEEE/CVF international conference on computer vision},
  pages={6023--6032},
  year={2019}
}

@article{pascalVOC,
  title={The pascal visual object classes challenge: A retrospective},
  author={Everingham, Mark and Eslami, SM Ali and Van Gool, Luc and Williams, Christopher KI and Winn, John and Zisserman, Andrew},
  journal={International journal of computer vision},
  volume={111},
  pages={98--136},
  year={2015},
  publisher={Springer}
}

@article{meanteacher,
  title={Mean teachers are better role models: Weight-averaged consistency targets improve semi-supervised deep learning results},
  author={Tarvainen, Antti and Valpola, Harri},
  journal={Advances in neural information processing systems},
  volume={30},
  year={2017}
}

@inproceedings{hu2023pseudo,
  title={Pseudo-label alignment for semi-supervised instance segmentation},
  author={Hu, Jie and Chen, Chen and Cao, Liujuan and Zhang, Shengchuan and Shu, Annan and Jiang, Guannan and Ji, Rongrong},
  booktitle={Proceedings of the IEEE/CVF international conference on computer vision},
  pages={16337--16347},
  year={2023}
}

@inproceedings{li2024semantic,
  title={Semantic consistency regularization with large language models for semi-supervised sentiment analysis},
  author={Li, Kunrong and Liu, Xinyu and Chen, Zhen},
  booktitle={International Conference on Neural Information Processing},
  pages={289--303},
  year={2024},
  organization={Springer}
}

@article{kumari2024leveraging,
  title={Leveraging task-specific knowledge from LLM for semi-supervised 3D medical image segmentation},
  author={Kumari, Suruchi and Das, Aryan and Roy, Swalpa Kumar and Joshi, Indu and Singh, Pravendra},
  journal={arXiv preprint arXiv:2407.05088},
  year={2024}
}

@article{mumcu2025leveraging,
  title={Leveraging Multimodal LLM Descriptions of Activity for Explainable Semi-Supervised Video Anomaly Detection},
  author={Mumcu, Furkan and Jones, Michael J and Cherian, Anoop and Yilmaz, Yasin},
  journal={arXiv preprint arXiv:2510.14896},
  year={2025}
}

@article{zheng2021rectifying,
  title={Rectifying pseudo label learning via uncertainty estimation for domain adaptive semantic segmentation},
  author={Zheng, Zhedong and Yang, Yi},
  journal={International Journal of Computer Vision},
  volume={129},
  number={4},
  pages={1106--1120},
  year={2021},
  publisher={Springer}
}



\end{document}